\documentclass[lettersize,journal]{IEEEtran}
\usepackage{amsmath,amsfonts}
\usepackage{algorithmic}
\usepackage{algorithm}
\usepackage{array}
\usepackage[caption=false,font=normalsize,labelfont=sf,textfont=sf]{subfig}
\usepackage{textcomp}
\usepackage{stfloats}
\usepackage{url}
\usepackage{verbatim}
\usepackage{graphicx}
\usepackage{cite}
\usepackage{booktabs}
\usepackage{xcolor}
\usepackage[many]{tcolorbox}
\usepackage{multirow}
\begin{document}

\title{Unveiling Deepfakes: A Frequency-Aware Triple Branch Network for Deepfake Detection}
\author{
Qihao Shen,
Jiaxing Xuan, 
Zhenguang Liu, 
Sifan Wu,
Yutong Xie, 
Zhaoyan Ming, 
Yingying Jiao, 
and kui Ren
\thanks{Qihao Shen, Jiaxing Xuan, Zhenguang Liu, and kui Ren are with the State Key Laboratory of Blockchain and Data Security, Zhejiang University, Hangzhou, China (e-mail: 77195039sqh@gmail.com; xuanjiaxing@sgdt.sgcc.com.cn; liuzhenguang2008@gmail.com; kuiren@zju.edu.cn).}
\thanks{Sifan Wu is with the College of Computer Science and Technology, Jilin University, Changchun 130012, China (e-mail: wusifan2021@gmail.com).}
\thanks{Jiaxing Xuan and Yutong Xie are also with State Grid Blockchain Technology (Beijing) Co., Ltd., Beijing, China (e-mail: xuanjiaxing@sgdt.sgcc.com.cn; xieyutong@sgdt.sgcc.com.cn).}
\thanks{Zhenguang Liu and Kui Ren are also with Hangzhou High-Tech Zone (Binjiang) Institute of Blockchain and Data Security.}
\thanks{Zhaoyan Ming is with Hangzhou City University, Hangzhou, China (e-mail: mingzhaoyan@gmail.com).}
\thanks{Yingying Jiao is with the College of Computer Science and Technology, Zhejiang University of Technology, Hangzhou, China (e-mail: yingyingjiao21@gmail.com).}
\thanks{Corresponding author: Zhenguang Liu.}
}

\IEEEpubid{0000--0000/00\$00.00~\copyright~2021 IEEE}

\maketitle

\begin{abstract}
Advanced deepfake technologies are blurring the lines between real and fake, presenting both revolutionary opportunities and alarming threats. While it unlocks novel applications in fields like entertainment and education, its malicious use has sparked urgent ethical and societal concerns ranging from \emph{identity theft} to the \emph{dissemination of misinformation}. To tackle these challenges, feature analysis using frequency features has emerged as a promising direction for deepfake detection. However, one aspect that has been overlooked so far is that existing methods tend to concentrate on one or a few specific frequency domains, which risks overfitting to particular artifacts and significantly undermines their robustness when facing diverse forgery patterns. %or cross-domain scenarios.%
Another underexplored aspect we observe is that different features often attend to the same forged region, resulting in redundant feature representations and limiting the diversity of the extracted clues. This may undermine the ability of a model to capture complementary information across different facets, thereby compromising its generalization capability to diverse manipulations.

In this paper, we seek to tackle these challenges from
two aspects: (1) we propose a triple-branch network that jointly captures spatial and frequency features by learning from both original image and image reconstructed by different frequency channels, and (2) we mathematically derive feature decoupling and fusion losses grounded in the mutual information theory, which enhances the model to focus on task-relevant features across the original image and the image reconstructed by different frequency channels. Extensive experiments on \emph{six} large-scale benchmark datasets demonstrate that our method consistently achieves state-of-the-art performance. Our code is released at https://github.com/injooker/Unveiling\_Deepfake. 
\end{abstract}


\section{Introduction}
\IEEEPARstart{T}{he} advancement of deep learning has propelled the growth of deepfake technology\cite{thies2016face2face,chen2020simswap,nirkin2019fsgan}, which allows the effortless creation of highly realistic fake videos that are nearly indistinguishable by the human eye\cite{liu2021deep,tolosana2020deepfakes,wang2023bi,chen2024sfe,yin2022mix,pan2023dfil}. 
While deepfake technology holds promise in benign domains like \emph{entertainment} and \emph{education}, its potential for malicious use has raised profound societal and ethical issues\cite{liu2022copy,suwajanakorn2017synthesizing}, such as \emph{identity fraud}, \emph{misinformation dissemination}, and \emph{privacy infringement}. A notable example occurred in March 2022, when cyber attackers released a fake video of Ukrainian President Zelensky urging soldiers to surrender~\cite{zelensky_deepfake}, leading to widespread confusion and mistrust. Such incidents underscore the urgent need for robust deepfake detection systems to combat these growing threats and rebuild public trust in digital media.

% 放在第一页第二栏的开头
\IEEEpubidadjcol

\begin{figure*}[htbp]
  \centering
  \includegraphics[width=0.8\textwidth]{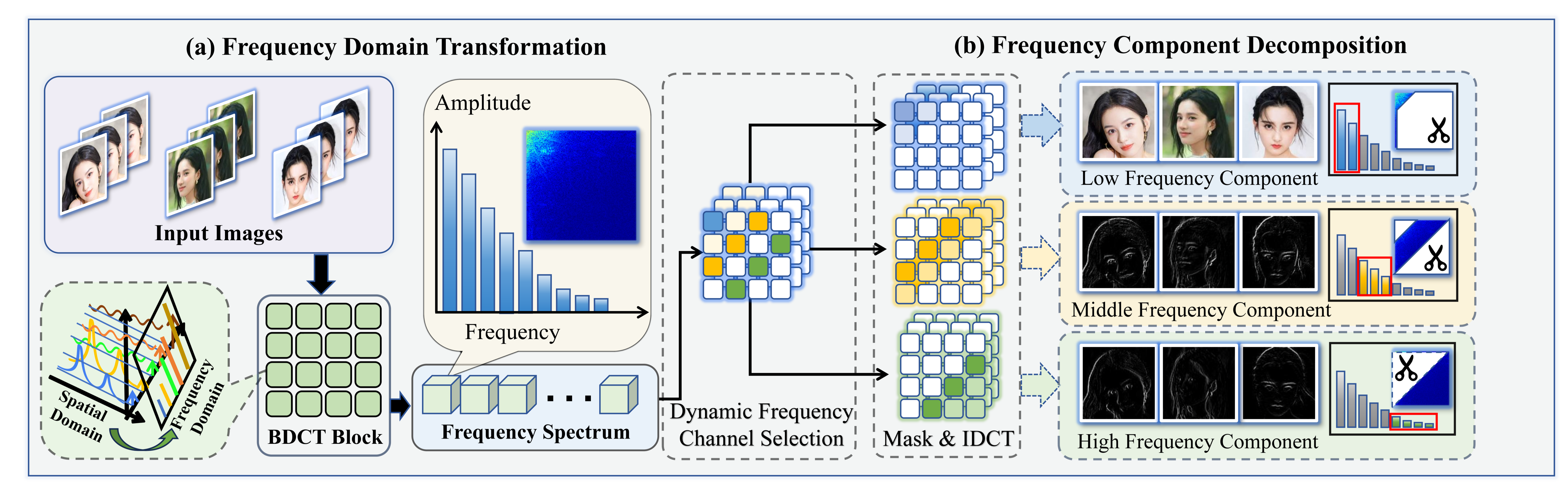} 
  \caption{The input image is transformed into the frequency domain via discrete cosine transform (DCT), where different frequency channels reveal distinct and complementary forgery cues. This observation motivates our dynamic frequency channel selection and cross-frequency enhancement, which enable comprehensive exploitation of frequency-domain forgery artifacts.}
  \label{fig:dct} 
\end{figure*}

 Despite significant progress in deepfake detection, existing methods still face two persistent and often overlooked challenges: (1) Most deepfake detection methods have concentrated on spatially noticeable forgery regions, such as \emph{mouth distortions}\cite{haliassos2021lips} and \emph{facial boundaries blur}\cite{li2020face}. However, as current forgery algorithms have become increasingly sophisticated in manipulating these areas, the accuracy of these detection methods shows a significant decline. (2) Current frequency-based approaches heavily rely on hand-engineered frequency bands, as shown in Fig.\ref{fig:dct}, which are often restricted to a limited number of pre-defined frequency channels. More importantly, current methods lack theoretical constraints guarantee their capacity to comprehensively exploit forgery-related information across spatial and frequency features\cite{qian2020thinking,masi2020two,fridrich2012rich}. These limitations significantly undermine their effectiveness to tackle the complexity and fast-paced evolution of deepfake detection. (3) To overcome these challenges, recent research has increasingly focused on spatial–frequency fusion approaches \cite{kou2025sfiad,zheng2025spatio,wang2024spatial}, which aim to leverage the complementary strengths of spatial and spectral representations. By integrating spatial cues with frequency-domain features, these methods can capture more subtle forgery traces and exploit cross-domain correlations that single-domain approaches often miss. However, existing fusion strategies generally rely on fixed or heuristic frequency channels and lack principled mechanisms to ensure complementary information across branches, such as dynamic frequency selection or mutual-information-driven guidance.

In light of these insights, we design a triple-branch network that jointly captures spatial-domain and frequency-domain features, enabling more robust detection of diverse forgery artifacts. Further, we introduce a dynamic frequency channel selection module driven by a lightweight attention, allowing the model to dynamically select channels into two frequency branches while minimizing irrelevant or redundant information. Then, we propose a cross-frequency enhancement module to merge the two frequency branches. To enhance feature diversity and reduce redundancy, we mathematically derive mutual information-driven losses that encourage the RGB branch and the merged frequency branch to focus on distinct yet complementary forgery features.

Empirically, we evaluate the proposed method on \emph{six} large scale benchmark datasets: DF40\cite{yan2024df40nextgenerationdeepfakedetection}, FaceForensics++\cite{rossler2019faceforensics++}, Celeb-DF\cite{li2020celeb}, Celeb-DF-v2, DFDC\cite{Deepfake-detection-challenge}, and DFDCP. Extensive experiments demonstrate that our approach achieves state-of-the-art performance in both in-dataset (training and testing on the same domain) and cross-dataset (training and testing on different domains) settings. Our method consistently and significantly outperforms existing approaches in terms of AUC on six datasets, highlighting its effectiveness in capturing forgery-related artifacts.

\textbf{Contributions.} In summary, our contributions are as follows:
\begin{itemize}
    \item We systematically investigate the joint exploitation of spatial and frequency domains to capture complementary and subtle forgery artifacts that are difficult to detect using either domain alone.
    \item We mathematically derive feature decoupling objectives based on mutual information theory to encourage feature diversity, and derive global information alignment objectives to effectively aggregate spatial and frequency features.
    \item We design a triple-branch network, which guides the model to extract spatial and frequency features across different branches. By reducing redundancy and diversifying learned features, this architecture demonstrates enhanced capability in detecting diverse forgery artifacts. Our code is released at https://github.com/injooker/Unveiling\_Deepfake to facilitate future research.
\end{itemize}

\section{Related Work}
The rapid advancement of deepfake technology\cite{shuai2023locate,pan2023dfil,li2024freqblender}, fueled by advances in deep learning and abundant large-scale datasets, has enabled the generation of highly realistic and indistinguishable synthetic media. While this technological progress demonstrates remarkable achievements in generative models, it simultaneously poses significant challenges for detection methods, demanding more multi-perspective and generalizable solutions. Existing deepfake detection approaches are broadly categorized into spatial-based and frequency-based approaches, each with unique strengths and inherent drawbacks.

\subsection{Spatial-Based Forgery Detection}

 Spatial-based detection methods operate in the pixel domain, identifying forgery through visual artifacts and semantic inconsistencies, such as anomalous facial features (e.g., irregular eye blinks, lip-sync discrepancies) or boundary distortions in manipulated regions\cite{dong2022protecting,fei2022learning,sun2022dual,zhuang2022uia}. Face X-ray \cite{li2020face} and SBIs \cite{shiohara2022detecting} detect forgery by analyzing the blending boundaries in face-swapped images, while methods like ICT \cite{dong2022protecting} employ augmented datasets to enhance discrimination between authentic and forged facial regions. Despite these methods have demonstrated effectiveness in detecting specific types of forgeries (e.g., face swapping), they are heavily reliant on pixel-level inconsistencies and manipulation-specific artifacts, which makes their performance tightly coupled with techniques encountered during training. As a result, they often fail to generalize to novel or evolving manipulation strategies and cross-dataset scenarios. Moreover, the localized analysis paradigm (e.g., focusing on eyes/mouth regions) may fail to capture globally distributed artifacts. Additionally, their reliance on visible artifacts becomes progressively less effective as generation techniques advance and visual imperfections become more subtle. To tackle these limitations, we propose a triple-branch network that jointly captures spatial and frequency features, enabling the detection of both localized and globally distributed artifacts.

\subsection{Frequency-Based Forgery Detection}
Frequency domain analysis has emerged as a powerful tool for deepfake detection, as it can reveal subtle artifacts that are often invisible in the spatial domain. Recent works have explored the use of frequency cues to identify forgeries \cite{li2021frequency}. For example, some methods \cite{frank2020leveraging} design filters to capture forgery clues at specific frequencies (e.g., high-frequency noise or low-frequency inconsistencies); others extract frequency domain features using techniques like Discrete Cosine Transform (DCT) or Fourier Transform to enhance detection performance \cite{qin2021fcanet}. Despite their demonstrated effectiveness, current frequency-based detection methods exhibit a critical limitation that most approaches concentrate their analysis on isolated frequency bands (e.g., exclusively examining either high-frequency or low-frequency components), while neglecting the potentially discriminative information contained in the complete spectral profile \cite{qian2020thinking}. More critically, existing approaches frequently depend on manually designed frequency filters or strong prior assumptions about the spectral distribution of forgery artifacts \cite{Durall_2020_CVPR}. While they incorporate frequency domain features, they do not fully exploit the distribution of forgery clues across all frequency channels. Furthermore, due to the absence of rigorous theoretical constraints, existing methods are prone to capturing redundant or non-forgery-related features \cite{li2024freqblender}.

%1.26增补
\subsection{spatial–frequency fusion Forgery Detection}

While spatial and frequency-based detection methods have proven effective independently, hybrid approaches that combine the strengths of both domains are gaining increasing attention.
Wang et al. [20] proposed a spatial–frequency feature fusion framework based on knowledge distillation, in which frequency-domain representations are leveraged to guide spatial feature learning, significantly enhancing robustness against compression and low-quality degradations. A Spatio-Frequency Cross Fusion Model [19] further introduces cross-domain interaction mechanisms to explicitly model the correlations between spatial and spectral features, enabling more effective exploitation of complementary forgery cues. More recently, SFIAD [18] integrates spatial–frequency feature fusion with dynamic margin optimization, aiming to improve both detection accuracy and decision boundary discrimination.

Despite these promising advances, existing methods still largely rely on heuristic fusion strategies or shallow interactions, without principled theoretical guidance to constrain cross-frequency feature integration and spatial–frequency\\ interactions. Consequently, their ability to fully leverage the complementary information of spatial and frequency domains is limited, especially when generalizing across datasets or challenging real-world conditions.

\section{Method}
\begin{figure*}[t] % [t] 强制图片浮动到顶部
  \centering % 图片居中
  \includegraphics[width=0.8\textwidth]{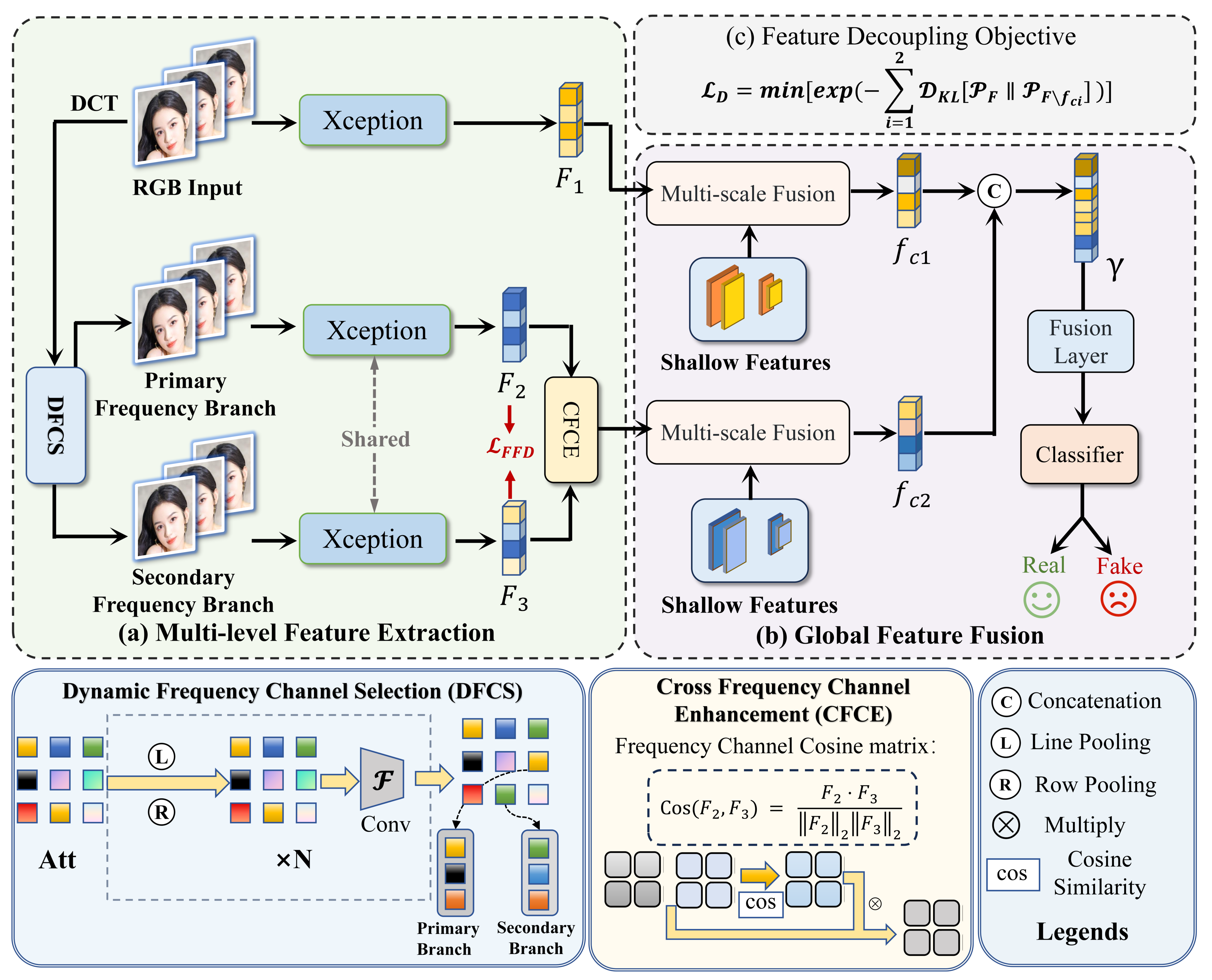} % 图片宽度设为 textwidth
  \caption{Overview of the proposed framework. The dynamic frequency channel selection (DFCS) module first transforms the input image into multiple frequency representations, providing complementary frequency-domain inputs beyond the RGB branch. The proposed cross frequency channel enhancement (CFCE) module further strengthens informative frequency cues by adaptively enhancing complementary information across frequency channels. To ensure that the merged feature and the feature extracted by the RGB branch are distinct from each other, a feature decoupling module (FDM) is introduced. Finally, the global fusion module (GFM) integrates multi-scale features while suppressing label-irrelevant and redundant information, yielding a compact and discriminative representation for reliable deepfake detection. Together, these components enable comprehensive exploitation of spatial–frequency forgery artifacts and constitute the core novelties of our approach.} % 图片描述
  \label{fig:framework} % 标签，用于引用
\end{figure*}

We empirically observe that despite the significant progress achieved by current deepfake detection methods, two key limitations remain. (1) Existing methods often heavily rely on detecting pixel-level inconsistencies and spatial anomalies, which are stuck to specific manipulation techniques. Although effective on seen datasets, existing methods usually fall short in generalizing to unseen manipulation types and cross-dataset scenarios. (2) Moreover, due to the lack of rigorous theoretical constraints, current methods may capture redundant or non-forgery-related features, leading to  sub-optimal detection performance.

To tackle these limitations, we advocate explorations in two aspects. (1) We propose a novel architecture that jointly captures spatial-domain and frequency-domain features for more robust deepfake detection. Our framework adopts a triple-branch design and integrates two key components: a \emph{dynamic frequency-channel selection} module and a \emph{cross frequency-channel enhancement} module, which enable the network to adaptively focus on informative frequency cues and capture complementary forgery features across different domains, thereby improving robustness to diverse and unseen manipulation patterns. (2) To mathematically constrain distinct feature learning, we propose a feature decoupling module by deriving mutual information losses to guide the RGB and frequency branches toward learning distinct forgery cues. Further, we design a global fusion module to integrate distinct features into a global feature. This dual-step design enables more accurate and comprehensive recognition of diverse manipulation features.

More specifically, our network consists of three specialized branches, namely \emph{RGB}, \emph{primary frequency}, and \emph{secondary frequency} branches. (i) The RGB branch focuses on extracting features directly from the input RGB image, preserving rich spatial and texture information. (ii) The primary frequency branch extracts features from the   image reconstructed by inverse discrete cosine transform  using only the top-K frequency channels. This serves to emphasize the most discriminative spectral components relevant to forgery detection. (iii) The secondary frequency branch extracts features from the image reconstructed by inverse discrete cosine transform using only the frequency channels ranked between K+1 and 2K. This serves to capture complementary yet moderately informative spectral cues. This triple-branch design ensures a comprehensive representation that synergistically combines spatial details and multi-level frequency information, thereby enhancing the ability of network to detect diverse and subtle manipulation traces.

As shown in Fig.\ref{fig:framework}, the processing pipeline begins in \emph{dynamic frequency-channel selection} module by transforming the input RGB image into its frequency representation using 8×8 block discrete cosine transformation ~\cite{xu2020learning,ji2022privacy}. The RGB branch processes the original image directly, retaining spatial textures crucial for visual artifact analysis. Simultaneously, the primary frequency branch utilizes the top-K most informative frequency channels to reconstruct the RGB image using inverse discrete cosine transform. Then we feed the reconstructed image into an Xception-based encoder to extract discriminative spectral features. The secondary frequency branch utilizes the frequency channels ranked between K+1 and 2K to reconstruct the RGB image using inverse discrete cosine transform. The reconstructed image is subsequently passed through an Xception-based encoder to capture secondary spectral features. Thereafter, the primary and secondary frequency features are merged through a \emph{cross frequency-channel enhancement} module, which integrates complementary cues across the two frequency branches. To ensure the feature extracted by the RGB branch and the merged feature are distinct from each other, a \emph{feature decoupling module} is introduced. Finally, a \emph{global fusion module} serves to integrate and compress features from both the RGB and merged branches, yielding a unified global representation for final classification.

\subsection{Dynamic Frequency Channel Selection}

The frequency-domain representation offers a compact yet informative view of visual signals, revealing subtle artifacts that may not be easily captured in the spatial domain. Leveraging this insight, our network is designed as a triple-branch architecture: the first branch directly processes the input RGB image to retain spatial textures critical for artifact localization, while the second and third branches focus on extracting spectral features from frequency-transformed inputs. Central to this design is the proposed \emph{dynamic frequency channel selection} module, which hierarchically identifies informative frequency channels from the image. Specifically, it selects the top-K most discriminative frequency channels for the second branch, and the next K frequency channels (ranked from K+1 to 2K) for the third branch, thereby enabling a fine-grained and complementary spectral analysis tailored for forgery detection.

Existing methods often rely on a few manually crafted and fixed low-frequency channels, based on the assumption that low-frequency components carry most of the information. However, our observations suggest that the most indicative frequency channels vary significantly across different images and manipulation types. Relying solely on fixed low-frequency channels risks overlooking valuable mid- or high-frequency artifacts. To overcome this limitation, the \emph{dynamic frequency channel selection} module automatically identifies and adapts to the most informative frequency channels. This flexibility allows our model to generalize better across diverse forgery scenarios.

 Our module dynamically identifies and prioritizes two distinct channel types: \emph{highly discriminative channels} containing concentrated manipulation artifacts, and \emph{moderately informative channels} with complementary forgery evidence. Through block discrete cosine transform, the input image $F_{0} \in \mathbb{R}^{c \times h \times w}$ is converted into its frequency representation $F \in \mathbb{R}^{c \times 64 \times \frac{h}{8} \times \frac{w}{8}}$. We propose a lightweight attention module that operates on globally pooled spectral features via two stacked 1×1 convolutions and a non-linear activation. Formally, for the frequency representation $F$, we first compute global average pooling ($\mathcal{G}$), then apply the lightweight attention module to generate a 64D channel attention vector, which is subsequently reshaped into an 8×8 importance map. Mathematically, the channel attention can be formulated as :
%  \begin{align}
% \mathbf{A}^* = 
% &\underbrace{
% \text{reshape} \left( 
% \mathcal{G}\left(F \right)
% \right)
% }_{\text{Channel-wise Attention}} \nonumber \\
% &\oplus
% \underbrace{
% \alpha \cdot \psi_\theta\left(
% \sum_{i=1}^{8} \mathcal{G}(F)_{i} \otimes 
% \sum_{j=1}^{8}  \mathcal{G}(F)_{j}^\top
% \right)
% }_{\text{Row-Column Attention (Conv $1{\times}1$)}},
% \label{eq:final_attention}
% \end{align}
\begin{align}
\mathbf{A}_{rc} = 
\underbrace{
\psi_\theta\left(
\sum_{i=1}^{8} \mathcal{G}(F)_{i} \otimes 
\sum_{j=1}^{8}  \mathcal{G}(F)_{j}^\top
\right)
}_{\text{Row-Column Attention (Conv $1{\times}1$)}},
\label{eq:final_attention}
\end{align}

correlations across the frequency domain. Specifically, $\sum_{i=1}^{8} \mathcal{G}(F)_i$ and $\sum_{j=1}^{8} \mathcal{G}(F)_j^\top$ compute the row-wise and column-wise aggregated responses of the $8 \times 8$ frequency attention map. The lightweight module $\psi\theta(\cdot)$, implemented as two stacked $1 \times 1$ convolutions and a non-linear activation, further refines these correlations and produces a compact yet expressive attention map. In this way, $\mathbf{A}_{rc}$ enables trend-aware modulation, effectively emphasizing frequency regions exhibiting strong directional patterns.  Finally, the overall attention representation is formulated as:

\begin{align}
\mathbf{A}^* = 
\mathcal{G}\left(F \right)
&\oplus
\alpha \cdot \mathbf{A}_{rc}
,
\label{eq:final_attention}
\end{align}

where $\oplus$ represents the structural fusion of dual attention branches, with $\alpha \in [0, 1]$ serving as a tunable scaling parameter for row-column enhancement.

Subsequently, the primary frequency branch utilizes the top-$K$ most informative frequency channels to reconstruct the RGB image. Then we feed the reconstructed image into an Xception-based encoder to extract discriminative spectral features. Similarly, the secondary frequency branch utilizes the frequency channels ranked between $K$+1 and 2$K$ to reconstruct the RGB image. The resulting image is passed through an Xception-based encoder to capture secondary spectral features.

The number of selected frequency channels, denoted as $K$, is an important hyperparameter in our \emph{dynamic frequency channel selection} module. Based on extensive validation and ablation experiments, we empirically set $K$ = 8 in all experiments, which achieves an optimal trade-off between detection accuracy and computational efficiency. A detailed sensitivity analysis of $K$ is provided in Section IV.C (Ablation Study on Channel Number), where we systematically evaluate the impact of varying $K$ on both in-dataset and cross-dataset performance.

\subsection{Cross Frequency Channel Enhancement}
While the features extracted by the \emph{primary frequency branch} and \emph{secondary frequency branch} provide a tiered view of features, utilizing them independently may lead to fragmented representations and overlook inter-channel dependencies. To mitigate this, our cross-frequency channel enhancement module enables collaborative learning between the two branches.

Specifically, the module operates on feature maps of highly discriminative channels denoted as $F_{c_1} \in \mathbb{R}^{c \times h \times w}$ and moderately informative channels denoted as $F_{c_2} \in \mathbb{R}^{c \times h \times w}$. We compute a cross-channel cosine map by taking the inner product of each element. The resulting cosine map is then used to reweight and fuse the two sets of channels, producing a merged frequency branch that integrates forgery features from both primary and secondary frequency branches. Through this collaborative mechanism, the module learns richer and more discriminative frequency features while preserving independent and representative information from both branches. Notably, the differences between the two branches are particularly noticeable in channels with subtle discriminative cues, whereas a naive summation often results in overly similar representations and loss of complementary information. This design ensures that the merged frequency branch captures a more comprehensive spectrum of forgery artifacts.

\subsection{Feature Decoupling Module}\label{decouple}
After cross frequency channel enhancement, we obtain the features extracted by the RGB branch and the merged frequency branch. However, these two branches might contain redundant information due to overlapping responses to forgery traces. To reduce redundancy between the features of the two branches, we derive the \emph{feature decoupling module} based on mutual information theory. Formally, given two extracted features $f_{c1}, f_{c2}$ of the two branches, our decoupling loss function enhances forgery-relevant information in the joint representation $F_c = f_{c1} \oplus f_{c2}$ while reducing redundancy.

%%我们基于互信息推导Frequency Decoupling Module。 Next , we derived......,参考刘青宇的论文
Mutual information is a metric from information theory that measures how much knowing one variable reduces uncertainty about another. Formally, given two variables $x_1$ and $x_2$, mutual information $\mathcal{I}(x_1, x_2)$ between $x_1$ and $x_2$ quantifies the amount of their shared information, which is given by:
\begin{equation}
    \mathcal{I}(x_1, x_2) = \mathbf{E}_{p(x_1, x_2)}\log \frac{p(x_1, x_2)}{p(x_1)p(x_2)},
\end{equation}
where \( \mathbf{E} \) denotes the expectation operator, \( p(x_1, x_2) \) is the joint probability distribution of \( x_1 \) and \( x_2 \), while \( p(x_1) \) and \( p(x_2) \) represent their respective marginal distributions.

%%这一段短一点

Our goal is to minimize the redundancy between the two features $f_{c1}$ and $f_{c2}$, namely:
\begin{equation}
    \min \mathcal{I}(f_{c_1}; f_{c_2}).
\end{equation}

Let $y$ denote label-relevant information, $ \mathcal{I}(f_{c_1};f_{c_2})$ can be decomposed into:
\begin{equation}
    \mathcal{I}(f_{c_1};f_{c_2}) = 
    \underbrace{
        \mathcal{I}(f_{c_1};f_{c_2};y)
    }_{\text{label-relevant}} + 
    \underbrace{
        \mathcal{I}(f_{c_1};f_{c_2}|y)
    }_{\text{label-irrelevant}},
\end{equation}
where term $\mathcal{I}(f_{c_1};f_{c_2};y)$ represents label-relevant information and term $\mathcal{I}(f_{c_1};f_{c_2}|y)$ stands for label-irrelevant information. Term $\mathcal{I}(f_{c_1};f_{c_2}|y)$ represents label-irrelevant information, and the global information alignment loss (which will be introduced later in Subsection~\ref{global fusion}) tries to compress 
$\mathcal{I}(f_{c_1};f_{c_2}|y)$to zero. Therefore, we may simplify minimize $\mathcal{I}(f_{c_1}; f_{c_2})$ to minimize $\mathcal{I}(f_{c_1};f_{c_2};y)$.  

Applying the chain rule of mutual information, we have:
\begin{equation}
    \mathcal{I}(y;F_c) = \mathcal{I}(f_{c1};y|f_{c2}) + \mathcal{I}(f_{c2};y|f_{c1}) + \mathcal{I}(f_{c1};f_{c2};y).
\end{equation}
After features $F_c = f_{c1} \oplus f_{c2}$ are extracted, $\mathcal{I}(y;F_c)$ is invariant. Thus, minimize $\mathcal{I}(f_{c_1};f_{c_2};y)$ is equivalent to:

\begin{equation}
    \max \big( \mathcal{I}(f_{c1}; y|f_{c2}) + \mathcal{I}(f_{c2}; y|f_{c1}) \big).
\end{equation}

However, direct computation of mutual information presents significant challenges. Following Ba et al. \cite{ba2024exposing}, $\mathcal{I}(f_{c1}; y|f_{c2}) + \mathcal{I}(f_{c2}; y|f_{c1})$ in Eq.~(6) has a lower bound:

\begin{equation}
     \sum_{i = 1}^{2}\mathcal{I}(f_{ci};y|F \backslash f_{ci}) \ge \sum_{i = 1}^{2}\mathcal{D}_{KL}[\mathcal{P}_F||\mathcal{P}_{F_\backslash f_{ci}}],
\end{equation}
where $\mathcal{P}_F$ = $\mathcal{P} (y | F)$ denotes the
predicted distributions, and $D_{KL}$ is the Kullback-Leibler
(KL) divergence. Practically, we can try to maximize the lower bound $\sum_{i = 1}^{2}\mathcal{D}_{KL}[\mathcal{P}_F||\mathcal{P}_{F_\backslash f_{ci}}]$ instead of $ \sum_{i = 1}^{2}\mathcal{I}(f_{ci};y|F \backslash f_{ci})$. We choose to minimize the negetive exponential form of the lower bound, which is formulated as:

\begin{equation}
     \mathcal{L}_{D}=min[exp(- \sum_{i = 1}^{2}\mathcal{D}_{KL}[\mathcal{P}_F||\mathcal{P}_{F_\backslash f_{ci}}])].
\end{equation}

This loss enhances the RGB and the merged branches to capture distinct information.

%backbone 小一些，Fusion layer规则形状
\begin{figure*}[htbp] % figure环境用于让图片浮动排版，[htbp]表示图片可出现在这里(h)、顶部(t)、底部(b)或单独成页(p)
  \centering % 使图片居中
  \includegraphics[width=0.8\textwidth]{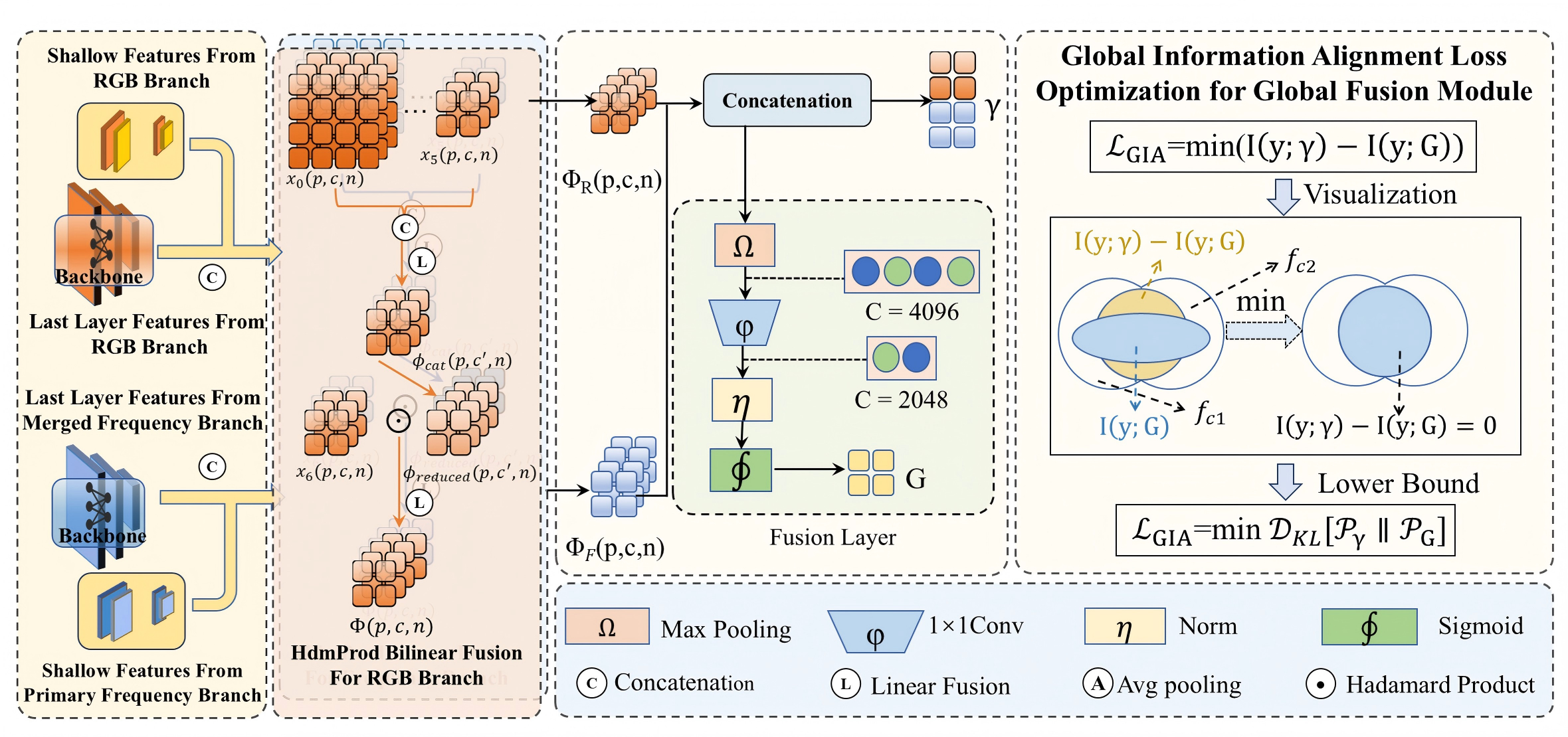} % 插入图片，假设图片名为example.jpg，这里将宽度设为5cm，可按需更改
   \caption{Detailed flowchart of the Global Fusion Module (GFM). The module transforms the simply concatenated features F into a discriminative fused representation G through multi-scale fusion and global information alignment, effectively reducing redundancy and enhancing label-relevant forgery cues.} % 图片的整体描述，会在图片下方显示
  \label{fig:global} % 给图片添加标签，方便后续交叉引用，例如 \ref{fig:example} 可引用该图片编号
\end{figure*}

\subsection{Global Fusion Module}\label{global fusion}
The feature decoupling module produces distinct forgery features. However, these features might contain label-irrelevant information. In the global fusion module, we introduce a multi-scale fusion strategy to integrate forgery features at different scales and design a global information alignment loss to remove label-irrelevant information. 

Many detection methods overlook shallow-layer forgery artifacts that are visually evident. To better exploit these discriminative shallow features, our global fusion module incorporates multi-scale fusion. The network employs Xception as backbone to extract hierarchical features from the RGB branch and the merged frequency branch.

We leverage features from Xception layers \(x_0\) to \(x_5\), where each \(x_i(p, c, n) \in \mathbb{R}^{H_i \times W_i \times C_i}\) represents the \(i\)-th layer output at spatial position \(p\), channel \(c\), and sample \(n\). These multi-scale features are concatenated channel-wise:

\begin{equation}
\phi(p, c, n) \overset{\text{cat}}{\longrightarrow} \bigoplus_{i=0}^{5} x_i(p, c, n)= 
\phi_{cat}(p, c', n),
\label{eq:concat}
\end{equation}
where \( \bigoplus \) represents the channel-wise concatenation operator. Features extracted from the layer \( \phi(p, c, n) \in \mathbb{R}^{H \times W \times C'} \) aggregates the multi-scale information, with \( C' = \sum_{i=0}^{5} C_i \).
To compress the high-dimensional concatenated features, we apply a \(1 \times 1\) depthwise separable convolution, generating a compact yet discriminative feature:

\begin{equation}
\phi_{\text{cat}}(p, c', n) 
\xrightarrow[]{1\times1\ \text{sep-conv}} 
\phi_{\text{reduced}}(p, c, n).
\label{eq:reduction}
\end{equation}

In this formulation, \( \phi_{\text{cat}} \) is the concatenated feature map, \( c' \in \mathbb{R}^{C'} \), and \( \phi_{\text{reduced}} \in \mathbb{R}^{H \times W \times C} \) is the reduced feature map after dimensionality reduction.

Finally, we integrate the compressed multi-scale features \(\phi_{\text{reduced}}\) with high-level semantic features \(x_6(p, c, n)\) from the final Xception layer through linear fusion (\(\mathcal{F}_{\text{LF}}\)), producing original fused features:

\begin{equation}
\phi_{\text{reduced}}(p, c, n),\ x_6(p, c, n) 
\overset{\mathcal{F}_{\text{LF}}}{\longrightarrow} 
\Phi(p, c, n),
\label{eq:hbfusion}
\end{equation}
where \( \Phi(p, c, n) \) represents the original fused feature map that integrates both hierarchical detail and global semantics.

To minimize the label-irrelevant information, namely $\mathcal{I}(f_{c_1};f_{c_2}|y)$ mentioned in \ref{decouple}, we design a global information alignment loss derived from mutual information theory. Our global fusion module processes original fused features $\gamma =\Phi(p, c, n)$ to eliminate label-irrelevant information through a dimensional reduction via Conv2D, and gets the compressed fused features $G$. As illustrated in Fig. \ref{fig:global}, minimizing $\mathcal{I}(f_{c_1};f_{c_2}|y)$ is equivalent to minimizing the difference between $\mathcal{I}(y;\gamma)$ and $\mathcal{I}(y;G)$:
\begin{equation}
     min\mathcal{I}(y;\gamma)-\mathcal{I}(y;G).
\end{equation}

We adopt the variational mutual information formulation from \cite{ba2024exposing}:
\begin{equation}
     min\mathcal{I}(y;\gamma)-\mathcal{I}(y;G) \Leftrightarrow min\mathcal{D}_{KL}[\mathcal{P}_\gamma||\mathcal{P}_{G}].
\end{equation}

From this, we can get the global information alignment loss:
\begin{equation}
     \mathcal{L}_{GIA}=min\mathcal{D}_{KL}[\mathcal{P}_\gamma||\mathcal{P}_{G}].
\end{equation}

Finally, the loss for our framework consists of a cross-entropy classification loss, a decoupling loss, and a global information alignment loss:
\begin{equation}
    \mathcal{L}_{total} = \mathcal{L}_{\text{CE}} + \alpha \mathcal{L}_{\text{D}} + \beta \mathcal{L}_{\text{GIA}},
\end{equation}
where $\alpha$ and $\beta$ are hyperparameters for the model.

%从去irrelevant的角度代替complementary

Our triple-branch network and associated loss functions effectively obtain distinct and label-relevant features from the RGB branch and the merged branch. The joint optimization of cross-entropy loss, decoupling loss, and global information loss ensures consistent performance across diverse datasets and manipulation methods.

\section{EXPERIMENTS}

\subsection{Settings}
\textbf{Implementation Details.} To ensure full reproducibility, we explicitly describe all experimental settings, including data preprocessing, training configurations, and optimization strategies. For data preprocessing, official annotations are aligned with the source videos, and RetinaFace is employed for accurate facial region localization and cropping, enabling focused analysis of forgery-relevant areas. All facial crops are resized to 299×299 resolution and normalized to the range [0, 1]. To enhance robustness and generalization, data augmentation strategies including random cropping, pixel contrast variation, Gaussian blurring, rotation, and grayscale conversion are applied while preserving annotation-image alignment. For model training, Xception initialized with ImageNet pre-trained weights is adopted as the backbone network. All experiments are conducted using the Adam optimizer \cite{kingma2014adam} with $\beta_1=0.9$, $\beta_2=0.999$, and $\epsilon=1\times10^{-8}$. The initial learning rate is set to $5\times10^{-4}$ and decayed by 0.5 every 5 epochs. The batch size is set to 64, and all models are trained for 20 epochs. The values of $\alpha$ and $\beta$ in Eq.(15) are automatically determined following the strategy proposed by Liebel and Körner \cite{liebel2018auxiliary}. All experiments are implemented using PyTorch on a workstation equipped with an NVIDIA RTX 3090 GPU (24GB memory).

\begin{table}[ht]
\centering
\small % 控制字体大小，不会太小
\caption{Details of the public datasets used in experiments.}
\label{tab:datasets}
\setlength{\tabcolsep}{4pt} % 减少列间距
\renewcommand{\arraystretch}{1.1} % 行间距略压缩
\begin{tabular}{lcccl}
\toprule
\textbf{Dataset} & \textbf{Real/Fake Videos} & \textbf{Year} & \textbf{Method} & \textbf{SSIM} \\
\midrule
DFDC-P      & 5,214        & 2019 & DeepFake        & 0.84 \\
DFDC        & 19,154/99,992& 2020 & DeepFake        & 0.84 \\
FF++        & 1,000/4,000  & 2019 & DF, F2F, FS, NT & 0.81 \\
CDF1        & 408/795      & 2020 & DeepFake        & 0.92 \\
CDF2        & 590/5,639    & 2020 & DeepFake        & 0.92 \\
DF40 \cite{yan2024df40nextgenerationdeepfakedetection}        & 10k+/0.1M+    & 2024 & FS,FR,EFS        & 0.92 \\
\bottomrule
\end{tabular}
\end{table}

\textbf{Datasets.} As shown in Table \ref{tab:datasets}, we evaluate our network on six public datasets, namely DF40 \cite{yan2024df40nextgenerationdeepfakedetection}, FaceForensics++ (FF++) \cite{rossler2019faceforensics++}, two versions of Celeb-
DF \cite{li2020celeb}, and two versions of Deepfake Detection
Challenge (DFDC) \cite{dolhansky2020deepfake} datasets. DFDC dataset includes DFDC-Preview (DFDC-P) and DFDC. DF40 comprises 40 distinct deepfake techniques, consists of 0.1M+ fake videos. DFDC-P as the preview of DFDC consists of
5,214 videos. DFDC, as one of the most large-scale face
swap datasets, contains more than 110,000 videos sourced
from 3,426 actors. FF++
dataset, which is the most widely used dataset, utilizes four
forgery-generation methods for producing 4,000 forgery
videos, $i.e.$, DeepFakes (DF)\cite{Deepfakes}, Face2Face (F2F)\cite{thies2016face2face}, FaceSwap
(FS)\cite{faceswap}, and NeuralTextures (NT)\cite{thies2019deferred}. FF++ has three compression versions and we use the high-quality level (C23) one
for training. Celeb-DF dataset has two versions, termed
Celeb-DF-V1 (CDF1) and Celeb-DF-V2 (CDF2). CDF1 consists of 408 pristine videos and 795 manipulated videos, while CDF2 contains 590 real videos and 5,639 deepFake videos.

\textbf{Evaluation Metrics.}
Our framework's performance is evaluated using three standard metrics: Accuracy (ACC), Area Under the Curve (AUC), and F1 score, which collectively measure classification performance across different aspects of real versus forged content discrimination.

% 表格标题
\begin{table*}[ht]
\centering
\setlength{\tabcolsep}{5pt} % 调整列间距
\renewcommand{\arraystretch}{1.2} % 调整行间距
\caption{In-dataset comparison results (frame-level AUC) on FF++ (C23) and Celeb-DF-V2 (CDF2).}
\label{tab:indataset}
\begin{tabular}{cc|cc}
\hline
\multicolumn{2}{c|}{\textbf{FF++(C23)}} & \multicolumn{2}{c}{\textbf{Celeb-DF-V2}} \\
\hline
\textbf{Method} & \textbf{AUC}$\uparrow$ & \textbf{Method} & \textbf{AUC}$\uparrow$ \\
\hline
Meso4 \cite{afchar2018mesonet} & 0.847 & Meso4 \cite{afchar2018mesonet} & 0.662 \\
Xception \cite{rossler2019faceforensics++} & 0.963 & DeepfakeUCL \cite{fung2021deepfakeucl} & 0.905 \\
Xception-ELA & 0.948 & SBIs(Shiohara et al. 2022) \cite{shiohara2022detecting} & 0.937 \\
SPSL \cite{masi2020two} & 0.943 & Agarwal et al. 2020\cite{agarwal2020detecting} & 0.990 \\
Face X-ray \cite{li2020face} & 0.874 & Wu et al. 2023\cite{wu2023generalizing} & 0.998 \\
TD-3DCNN \cite{zhang2021detecting} & 0.722 & TD-3DCNN\cite{zhang2021detecting} & 0.888 \\
F³-Net \cite{qian2020thinking} & 0.981 & Xception\cite{rossler2019faceforensics++} & 0.985 \\
FT-two-stream \cite{hu2021detecting} & 0.925 & FT-two-stream\cite{hu2021detecting} & 0.867 \\
FInfer \cite{hu2022finfer} & 0.957 & FInfer\cite{hu2022finfer} & 0.933 \\
Exposing the Deception \cite{ba2024exposing} & 0.983 & Exposing the Deception\cite{ba2024exposing} & 0.999 \\
FreqBlender \cite{li2024freqblender} & 0.981 & FreqBlender \cite{li2024freqblender} & 0.993 \\
\hline
\textbf{Ours (Xception)} & \textbf{0.990} & \textbf{Ours (Xception)} & \textbf{0.999} \\
\hline
\end{tabular}
\end{table*}
\subsection{Evaluations}
In this section, we benchmark our method against state-of-
the-art deepfake detection methods for in-dataset and cross-
dataset settings.

\textbf{In-dataset Performance.} For in-dataset evaluation, we conduct rigorous experiments on two standard benchmarks: FF++ (C23) and CDF2. As Table \ref{tab:datasets} shows, we compare with state-of-the-art methods using their originally reported results when source codes are unavailable, ensuring fair and consistent benchmarking. Table \ref{tab:indataset} demonstrates the consistent superiority of our method across all benchmarks. On FF++ (C23), we achieve a state-leading AUC of 0.990 (vs 0.987 previously). The performance gap widens on CDF2, where our 0.999 AUC surpasses Wu et al.'s \cite{wu2023generalizing} 0.998. These results validate the robustness of our network in in-distribution detection while advancing the performance standards. The consistent outperformance across both datasets underscores our method's practical viability for real-world deployment.

\textbf{Cross-dataset Performance.} Cross-dataset evaluation poses greater challenges by testing generalization across unseen distributions and manipulation techniques. We assess the robustness of our method using a strict strategy: training on FF++ (C23) and testing on five unseen datasets (CDF1, CDF2, DFDC-P, DFDC,DF40). As Table \ref{tab:crossdataset} shows, through frame-level AUC comparisons, this rigorous evaluation demonstrates our approach's superior generalizability compared to existing methods.

Our first observation is that state-of-the-art deepfake detection methods, despite their strong in-dataset performance, exhibit relatively low AUC scores on unseen datasets. This highlights a common limitation in the field: many methods tend to overfit to the specific artifacts and patterns present in the training dataset, resulting in poor generalization to new and diverse datasets. For example, methods like DCL \cite{sun2022dual} and UCF \cite{yan2023ucf} achieve AUCs of 0.767 and 0.719 on DFDC-P and DFDC, respectively, indicating their limited ability to adapt to unseen data distributions.

In contrast, our method demonstrates significantly greater robustness and generalization capability. When tested on unseen datasets, our approach achieves substantial improvements over existing methods. Specifically, our method attains frame-level AUCs of 0.796 and 0.872 on CDF1 and CDF2, respectively, outperforming the current state-of-the-art method DCL. On the DFDC-P dataset, our method improves the AUC from 0.767 (DCL) to 0.777, and on the DFDC dataset, it increases the AUC from 0.719 (UCF) to 0.735. These results underscore the effectiveness of our framework in addressing the overfitting problem and generalizing to diverse deepfake datasets.

Overall, our method achieves state-of-the-art generalization performance across multiple datasets, setting a new benchmark for cross-dataset deepfake detection. These results demonstrate the potential of our approach to serve as a reliable solution for real-world applications, where models must generalize to unseen and diverse deepfake manipulations.
% 表格标题
\begin{table*}[h]
\centering
\setlength{\tabcolsep}{5pt} % 调整列间距
\renewcommand{\arraystretch}{1.2} % 调整行间距

\caption{Cross-dataset comparison results (frame-level AUC) on Celeb-DF-V1 (CDF1), Celeb-DF-V2 (CDF2), DFDC-P, DFDC, and DF40.}
\label{tab:crossdataset}
\resizebox{\textwidth}{!}{
\begin{tabular}{lcccccccc}
\toprule
\textbf{Method}            & \textbf{Training dataset} & \textbf{CDF1} & \textbf{CDF2} & \textbf{DFDC-P} & \textbf{DFDC}& \textbf{FS(DF40-CDF)}& \textbf{FR(DF40-CDF)}& \textbf{EFS(DF40-CDF)} \\ 
\midrule
Xception \cite{rossler2019faceforensics++}    & FF++ & 0.750  & 0.778  & 0.698  & 0.636 & 0.922 & 0.657& 0.622 \\ 
DSP-FWA \cite{li2018exposing}  & FF++ & 0.785  & 0.814  & 0.595 & - & - & - & - \\ 
Meso4 \cite{afchar2018mesonet}  & FF++  & 0.422  & 0.536 & 0.594& -  & - & - & - \\ 
F$^3$-Net \cite{qian2020thinking}  & FF++  & -   & 0.712* & 0.729*  & 0.646* & - & - & - \\ 
SRM \cite{fridrich2012rich} & FF++ & 0.793 & 0.755& 0.741 & 0.700  & 0.919 & 0.621 & 0.603\\ 
CORE \cite{ni2022core}  & FF++  & 0.780*  & 0.743*  & 0.734*  & 0.705* & - & - & -\\ 
UCF \cite{yan2023ucf}     & FF++ & 0.779* & 0.753*  & 0.759*  & 0.719*  & - & - & - \\ 
RFM \cite{wang2021representativeforgeryminingfake}& FF++ & -& -   & 0.674  & 0.680* & \textbf{0.939} & 0.637 & 0.628 \\ 
HCIL \cite{gu2022hierarchical} & FF++ & - & 0.790 & 0.692 & -  & - & - & -\\ 
SPSL \cite{liu2021spatialphaseshallowlearningrethinking}  & FF++& 0.791  & 0.749 & 0.72& 0.696 & 0.938 & 0.646 & 0.648 \\ 
RECCE \cite{cao2022end}& FF++ & -  & 0.687  & - & 0.691& 0.926 & 0.632 & 0.610 \\ 
DCL \cite{sun2022dual}& FF++  & - & 0.823  & 0.767& -  & - & - & -\\
F-G \cite{lin2024preservingfairnessgeneralizationdeepfake}& FF++  & - & 0.744  & -& 0.617 & - & - & - \\
CFM \cite{luo2023priorforgeryknowledgemining}& FF++  & - & 0.828  & 0.758& - & - & - & - \\
\hline
\textbf{Ours (Xception)}& \textbf{FF++}& \textbf{0.796} & \textbf{0.872} & \textbf{0.777}  & \textbf{0.735}& 0.936 & \textbf{0.671}  & \textbf{0.656} \\ 
\bottomrule
\end{tabular}
}
\end{table*}

\subsection{Ablation Study}
\subsubsection{\textbf{Ablation Study of Triple-branch Network}}

\begin{table*}[h]
\centering
\setlength{\tabcolsep}{5pt} % 调整列间距
\renewcommand{\arraystretch}{1.2} % 调整行间距
\caption{Ablation study of different model components.}
\label{tab:ablation}
\begin{tabular}{c|c|cc|cc}
\hline
\textbf{Method} & \textbf{Training Dataset} & \multicolumn{2}{c|}{\textbf{FF++(C23)}} & \multicolumn{2}{c}{\textbf{CDF2}} \\ \cline{3-6} 
                &                       & \textbf{ACC}     & \textbf{AUC}     & \textbf{ACC}     & \textbf{AUC}     \\ \hline
RGB             & FF++                  & 0.945            & 0.966            & 0.742            & 0.813            \\ 
Frenquency Branch             & FF++    & 0.940            & 0.964            & 0.728            & 0.807            \\ 
Triple Branches      & FF++                  & 0.955           & 0.980            & 0.762            & 0.848            \\ 
Triple Branches+CFCE  & FF++                  & 0.958            & 0.983            & 0.775            & 0.857            \\ \hline

Triple Stream Network& FF++                  & \textbf{0.964}   & \textbf{0.988}   & \textbf{0.792}   & \textbf{0.872}   \\ \hline
\end{tabular}
\end{table*}

To rigorously validate the efficacy of individual components in our network, we conduct a systematic ablation study. This study quantitatively analyzes the impact of key modules and configurations on network performance. Specifically, we evaluate the following variants:
\begin{itemize}
    \item \textbf{RGB branch or frequency branch:} This establishes the baseline, where we independently assess the RGB branch or the frequency branch. This analysis isolates the contribution of each modality to the overall performance.
    \item \textbf{Triple Branches:} This variant obtains the two frequency branches from the \emph{dynamic frequency-channel selection} module (DFCS). Then, it combines the RGB branch and the two frequency branches through direct feature summation, enabling us to evaluate the advantages of naive multi-modal fusion without complex feature refinement.
    \item \textbf{Triple Branches + CFCE:} This variant incorporates our proposed \emph{dynamic frequency channel selection} module and the \emph{cross-frequency channel enhancement} (CFCE) module into feature extraction from the RGB branch and the merged branch.
    \item \textbf{Triple Stream Network:} Our complete framework integrates two key components of mutual information losses: the \emph{feature decoupline} (FDM) module, and \emph{the global fusion} (GFM) module.
\end{itemize}

Table \ref{tab:ablation} summarizes our experimental results, with all networks trained on FF++ and evaluated on both FF++ and CDF2 datasets. The key observations reveal: (1) the DFCS module achieves significant performance gains over single-modal baseline (RGB or frequency), demonstrating that multimodal integration effectively captures complementary information to enhance generalization capability. (2) The CFCE module yields consistent performance gains over the baseline triple-branch network, demonstrating its effectiveness in enforcing collaborative learning between the two branches. (3) The full network, which includes DFCS, CFCE, FDM, and GFM modules, achieves the best performance across both datasets. This suggests that the FDM and GFM modules play a crucial role in further refining the feature extraction. Specifically, the FDM helps in capturing distinct label-relevant information, while the GFM module enhances the global complementary features.

To highlight the key takeaways of the ablation results, we provide a concise summary with quantitative evidence. Among all components, the \textbf{DFCS module contributes the most to performance improvement}. Compared with the single-modality baselines, incorporating DFCS to form the triple-branch architecture improves the AUC on CDF2 from \textbf{0.813 (RGB)} and \textbf{0.807 (frequency)} to \textbf{0.848}, and the ACC on CDF2 from \textbf{0.742 (RGB) and 0.728 (frequency)} to \textbf{0.762}. This suggests that adaptive frequency-aware feature modeling and multimodal integration significantly enhance both \textbf{in-dataset performance and cross-dataset generalization}. In contrast, the \textbf{CFCE module mainly serves as a feature refinement mechanism}, yielding consistent but relatively moderate gains. Specifically, CFCE further improves the AUC on CDF2 from \textbf{0.848 to 0.857} and the ACC from \textbf{0.762 to 0.775}, indicating its effectiveness in refining multimodal feature interaction. Owing to its generic design, CFCE could potentially be replaced by other attention-based or feature fusion strategies without fundamentally altering the framework. Meanwhile, the \textbf{FDM and GFM modules act as auxiliary yet indispensable components}, which further boost the AUC to \textbf{0.872} and the ACC to \textbf{0.792} on CDF2, and improve the in-dataset performance on FF++ from \textbf{0.983 to 0.988} in terms of AUC. Although their individual improvements are less pronounced than those introduced by DFCS, they play a crucial role in stabilizing optimization, disentangling label-relevant information, and enhancing global feature aggregation. These empirical results confirm that \textbf{DFCS constitutes the core driving factor} of the proposed framework, while the remaining modules collectively contribute to the performance of the proposed framework.

In summary, our ablation study systematically validates the contribution of each component. The results conclusively show that integrating multi-modal fusion with cross-modal consistency enhancement and hierarchical feature refinement yields an optimally robust solution for this task.
\subsubsection{\textbf{Ablation Study on Channel Number}}

\begin{figure}[htbp] % [t] 强制图片浮动到顶部
  \centering % 图片居中
  \includegraphics[width=8cm]{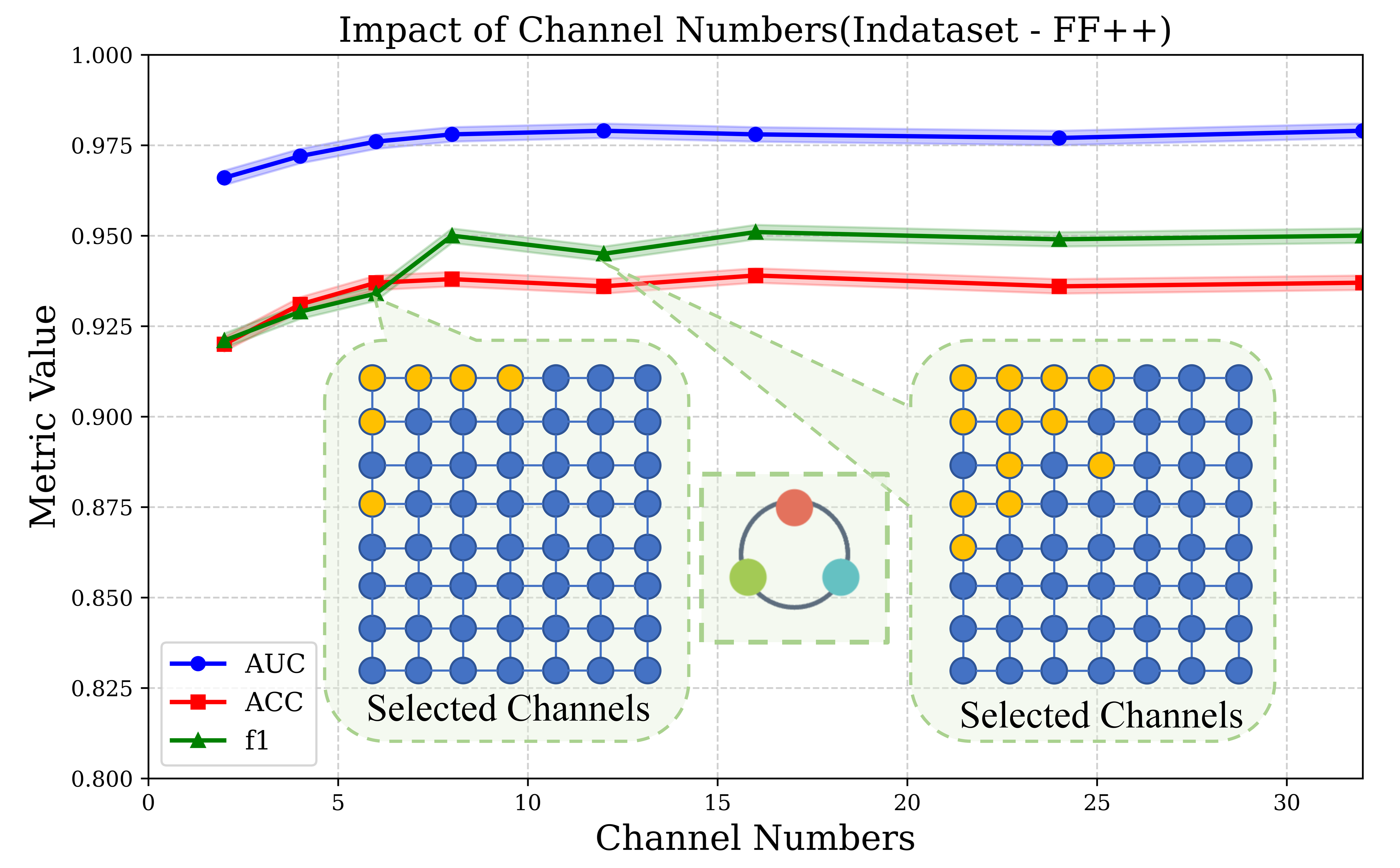} % 图片宽度设为 textwidth
  \caption{In-dataset impact of channel number on AUC, ACC and F1 score. The figure illustrates the relationship between the number of channels and the performance for deepfake detection on the FF++ dataset. As the number of channels increases, both AUC , ACC, F1 score show a trend of improvement. This demonstrates the importance of channel selection in optimizing model performance for forgery detection tasks.} % 图片描述
  
  \label{fig:channelNumber} % 标签，用于引用
\end{figure}

To systematically evaluate the effect of channel selection in our \emph{dynamic frequency channel selection} module, we conducted controlled ablation studies varying the key parameter $K$. All experiments were conducted with Xception as the backbone network, and all configurations were trained for 20 epochs using standardized data augmentation strategies (cropping, rotation, and grayscale conversion) to ensure experimental consistency.

Figure \ref{fig:channelNumber} demonstrates a strong correlation between channel selection and model performance. Severely constrained channel configurations ($K$=2/4) exhibit significantly degraded performance across all metrics (AUC, ACC, F1), indicating that inadequate channel capacity critically impairs feature extraction for classification tasks.

However, performance metrics saturate when exceeding a critical channel threshold ($K$\textgreater8), demonstrating that while sufficient channels are essential for peak performance, additional channels provide diminishing returns in AUC, ACC and F1. This insight enables optimal efficiency by balancing model performance with computational economy.

\begin{figure}[htbp] % [t] 强制图片浮动到顶部
  \centering % 图片居中
  \includegraphics[width=8cm]{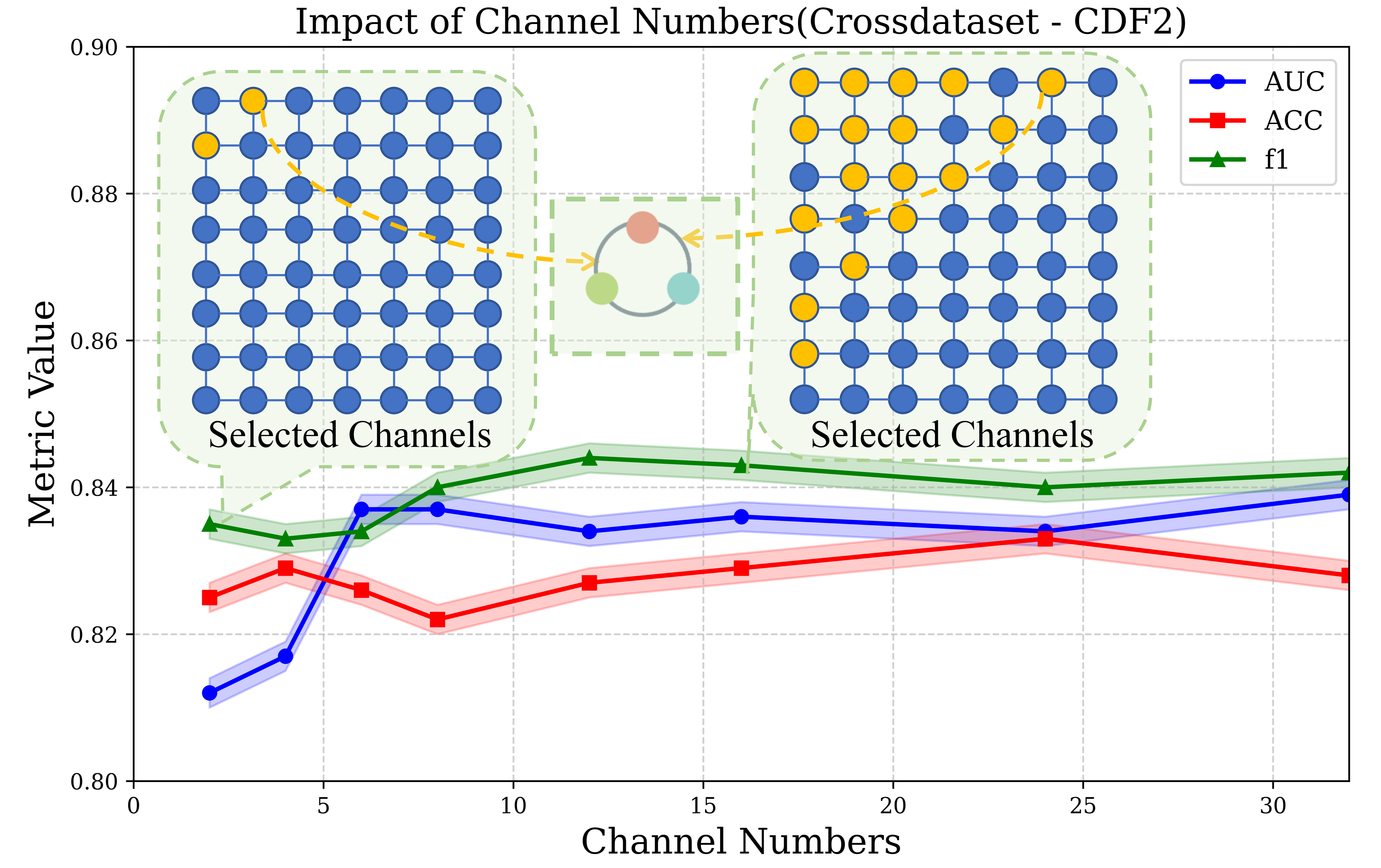} % 图片宽度设为textwidth
  \caption{Cross-dataset impact of channel number on AUC, ACC and F1 score. The figure illustrates the relationship between the number of channels and the performance metrics (AUC, ACC and F1 score) for deepfake detection on the CDF2 dataset. The results suggest that there is an optimal range for the number of frequency channels, where the model achieves the best trade-off between performance and computational efficiency.} % 图片描述
  \label{fig:channelNumberCross} % 标签，用于引用
\end{figure}

To systematically assess how frequency channel selection affects cross-dataset generalization, we conducted controlled experiments measuring AUC, ACC, and F1 on CDF2 while varying channels in our feature decoupling module. Figure \ref{fig:channelNumberCross} presents these results. Our frequency channel analysis reveals a nonlinear relationship between channel count and detection performance. The AUC shows monotonic improvement until reaching an optimal channel count ($K$=8), demonstrating that comprehensive frequency coverage is essential for robust forgery detection. However, we observe diminishing returns beyond this threshold, with marginal performance degradation at higher channel counts ($K$\textgreater16), suggesting channel redundancy may introduce spectral noise that compromises model generalization.

Consistent with AUC trends, both accuracy (ACC) and F1 score exhibit progressive improvement until reaching an optimal channel threshold, confirming that channel balance critically governs classification performance. Notably, ACC and F1 saturate earlier than AUC (at $K$=6 vs $K$=8), revealing their reduced sensitivity to additional channels once sufficient spectral coverage is achieved.
Our findings identify a precise operational window ($K$=6-8 channels) that optimally balances detection performance with computational efficiency. Below this range, insufficient spectral analysis compromises forgery detection capability; beyond it, diminishing returns transition into computational overhead without commensurate accuracy gains.

These findings provide critical guidance for efficient model design, especially in resource-constrained applications. Optimizing the channel selection enables an effective performance-efficiency trade-off, maximizing detection capability while minimizing computational overhead.
\subsection{Robustness}

\begin{table*}[h]
\small
\centering
\caption{Impact of image compression and noise on the model.}
\label{tab:robust}

\begin{tabular}{c|c|cc|cc}
\hline
\textbf{Method} & \textbf{Training set} & \multicolumn{2}{c|}{\textbf{FF++(C23)}} & \multicolumn{2}{c}{\textbf{CDF2}} \\ \cline{3-6} & & \textbf{ACC} & \textbf{AUC} & \textbf{ACC}     & \textbf{AUC}     \\ \hline
Original  & FF++   & 0.964  & 0.988  & 0.792  & 0.872 \\ 
Original+Image compression  & FF++   & 0.954 & 0.984  & 0.774 & 0.870  \\ 
Original+GaussNoise   & FF++  & 0.958  & 0.982  & 0.776 & 0.869 \\ 
Original+ISONoise  & FF++ & 0.956  & 0.981 & 0.782 & 0.865 \\ 
\hline
\end{tabular}

\end{table*}

In real-world scenarios, video degradation (e.g., low resolution and noise) often obscures forensic traces, significantly challenging deepfake detection. This section validates our model's robustness against such degradations, demonstrating its practical applicability under suboptimal conditions.

To tackle the challenge of low-resolution videos, our model employs a multi-scale feature extraction module that effectively captures forensic artifacts across varying resolutions. We further enhance robustness through targeted data augmentation, incorporating synthetically degraded low-resolution samples during training. For noise resilience, we systematically inject diverse noise patterns (e.g., Gaussian, ISO) into the training pipeline, enabling the model to learn noise-invariant detection features.

The experimental results after adding image compression and noise are shown in the table \ref{tab:robust}. The ability to handle low-resolution videos and noise variations is crucial for deploying deepfake detection systems in real-world applications, where video quality is often suboptimal. Our results demonstrate that data augmentation is effective strategies for improving model robustness. The demonstrated effectiveness of our targeted augmentation strategy confirms that training with synthetic distortions significantly enhances operational robustness, enabling reliable forgery detection in non-ideal conditions.

\subsection{Visualization}

\begin{figure*}[htbp] % [t] 强制图片浮动到顶部
  \centering % 图片居中
  \includegraphics[width=0.8\textwidth]{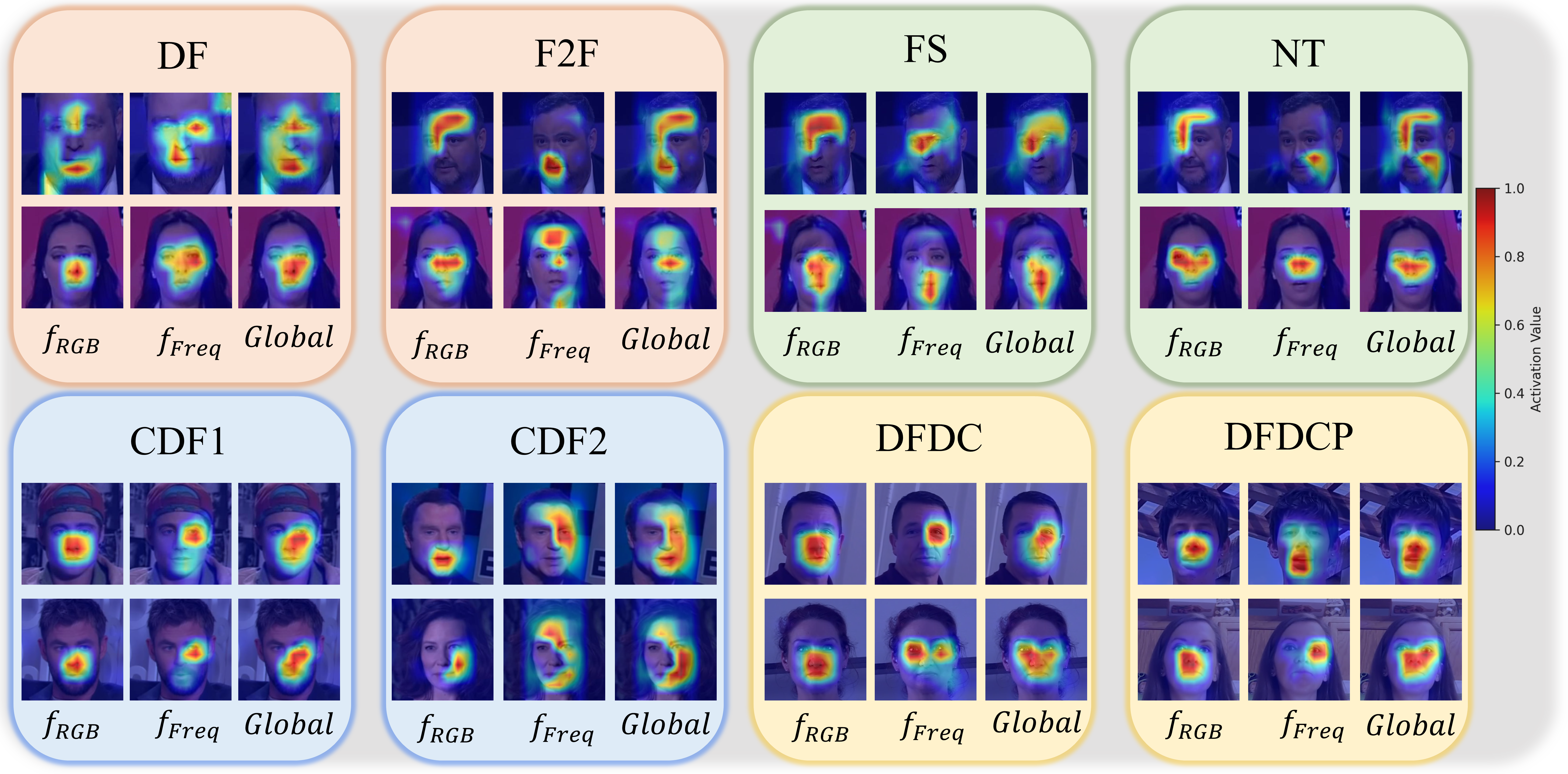} % 图片宽度设为 textwidth
  \caption{Visual examples of our method on Deepfakes (DF), Face2Face (F2F), FaceSwap (FS), NeuralTextures (NT), CDF, CDF2, DFDC and DFDC-P datasets. We can be observed that each branch focuses on distinct and complementary forgery-related regions, exhibiting minimal spatial overlap in their attention distributions. This behavior demonstrates the strong complementarity and functional specialization among branches, enabling the model to capture distinct forensic evidence across spatial and frequency domains, and thereby enhancing both detection robustness and generalization capability.} % 图片描述
  \label{fig:cam} % 标签，用于引用
\end{figure*}

To further evaluate the interpretability and effectiveness of our triple-branch network, we employ Gradient-weighted Class Activation Mapping (Grad-CAM) \cite{selvaraju2017grad} to visualize the regions of interest (ROIs) that the model focuses on when detecting deepfake content. Our evaluation encompasses eight representative samples from FF++ spanning all major manipulation categories (DF, F2F, FS, NT). As demonstrated in Figure \ref{fig:cam}, these visualizations reveal our model's ability to consistently localize authentic manipulation artifacts, and its superior focus on physiologically plausible regions compared to single-branch baselines.

Our analysis reveals that the proposed method exhibits comprehensive artifact localization capability, consistently identifying diverse forensic traces across all facial regions. In contrast to conventional approaches that focus narrowly on specific facial landmarks (e.g., periocular regions or mouth boundaries), our multi-modal framework effectively captures spatially distributed manipulation evidence. As evidenced in the deepfake samples, the model successfully detects both prominent boundary artifacts and subtle physiological inconsistencies (e.g., micro-texture anomalies and illumination discontinuities) that existing methods frequently miss.

The Grad-CAM visualizations demonstrate that our three branches exhibit specialized and complementary attention patterns, with remarkably low inter-branch overlap. This specialization enables each branch to contribute distinct forensic evidence. This hierarchical feature analysis provides coverage of manipulation traces across spatial scales, significantly outperforming single-branch approaches in detecting diverse deepfake variants.

Our method demonstrates a strong and consistent ability to localize forgery artifacts across diverse manipulation techniques. Specifically, on Face2Face samples, it accurately captures subtle yet telltale artifacts around key facial regions such as the mouth and eyes—areas frequently affected by reenactment-based manipulations. For NeuralTextures forgeries, the model successfully identifies nuanced inconsistencies in facial expressions and texture fidelity, underscoring its robustness against more sophisticated and photorealistic forgery approaches. These results collectively highlight the generalizability and precision of our method in tackling a wide range of deepfake scenarios.

\subsection{Limitations and Future Work}
\subsubsection{\textbf{Impact of Domain Shift and Robustness Analysis}}
Although our method demonstrates strong performance on cross-dataset evaluations, its generalization capability can still be affected by significant domain shifts, particularly low-level appearance variations such as resolution degradation and compression artifacts. These shifts directly distort the underlying pixel distribution and frequency statistics, posing substantial challenges for frequency-based detection models. In real-world scenarios, videos often undergo re-sampling, re-encoding, and platform-dependent compression, which makes such domain shifts especially critical.

\begin{table}[h]
\small
\centering
\caption{Effect of data augmentation on robustness against image degradation.}
\label{tab:robust_aug}
\resizebox{\columnwidth}{!}{
\begin{tabular}{c|c|cc|cc}
\hline
\textbf{Training set} & \textbf{Test degradation} 
& \multicolumn{2}{c|}{\textbf{FF++(C23)}} 
& \multicolumn{2}{c}{\textbf{CDF2}} \\ \cline{3-6}
 &  & \textbf{ACC} & \textbf{AUC} & \textbf{ACC} & \textbf{AUC} \\
\hline
Original & Compression & 0.938 & 0.972 & 0.762 & 0.850 \\
Original + Aug & Compression & \textbf{0.952} & \textbf{0.981} & \textbf{0.776} & \textbf{0.866} \\
\hline
Original & Gauss Noise & 0.942 & 0.975 & 0.768 & 0.858 \\
Original + Aug & Gauss Noise & \textbf{0.956} & \textbf{0.983} & \textbf{0.781} & \textbf{0.870} \\
\hline
Original & ISO Noise & 0.940 & 0.973 & 0.765 & 0.854 \\
Original + Aug & ISO Noise & \textbf{0.954} & \textbf{0.982} & \textbf{0.779} & \textbf{0.868} \\
\hline
\end{tabular}
}
\end{table}

To investigate whether simple data augmentation strategies can alleviate this issue, we conduct additional robustness experiments by incorporating image compression and noise injection during training, and evaluating the models under degraded test conditions, as shown in Table~VI. The results indicate that these augmentations can partially enhance robustness, consistently improving detection accuracy and AUC across different perturbation settings. However, the performance gains remain limited, especially on cross-dataset benchmarks, suggesting that simple augmentation alone is insufficient to fully bridge the domain gap caused by severe distribution mismatch.

From a frequency perspective, high-frequency components are affected first and suffered the most under resolution reduction and compression. Downsampling and lossy compression tend to suppress fine-grained high-frequency details that encode subtle forgery artifacts, while low-frequency structural information remains relatively stable. This explains the observed sensitivity of frequency-domain features to such domain shifts and motivates future exploration of more advanced domain generalization strategies to further improve robustness.
 
\begin{table}[h]
\small
\centering
\caption{Comparison of Model Parameters, GFLOPs and GPU time with other backbones.}
\label{tab:flops}

% \begin{tabular}{l|c|c}
% \hline
% \textbf{Model} & \textbf{Params(M)} & {\textbf{FLOPs(G)}} \\ \hline
% Xception & 20.82 & 9.69 \\ 
% VGG16 & 138.36 & 15.5 \\ 
% WideResNet101 & 126.89 & 22.84 \\ 
% PolyNet & 95.37 & 34.90 \\ \hline
% triple-branch network(ours) & 95.45  & 30.10 \\ \hline
% \end{tabular}
\begin{tabular}{l|c|c|c}
\hline
\textbf{Model} & \textbf{Params(M)} & \textbf{FLOPs(G)} & \textbf{GPU(ms)} \\
\hline
Xception & 20.82 & 9.69 & 6.39 \\
ResNet101 & 42.50 & 14.58 & 9.77 \\
Res2Net-101 & 43.16 & 15.08 & 13.91 \\
EfﬁcientNet-B7 & 63.79 & 10.20 & 21.11 \\
Swin-Base & 86.75 & 18.84 & 15.95 \\
WideResNet101 & 126.89 & 22.84 & 17.27\\ 
\hline
Ours & 95.45 & 30.10 & 22.37 \\
\hline
\end{tabular}
\end{table}

\subsubsection{\textbf{Computational Complexity and Model Efficiency}}
 As shown in Table \ref{tab:flops}, our model incorporates three parallel Xception backbones along with multi-scale fusion layers, resulting in approximately three times the computational cost compared to a single Xception network. Specifically, the number of parameters increases from 20.82M to 95.45M FLOPs, raising the training and inference overhead. To measure inference time, we used the thop library to compute FLOPs and conducted experiments on 300 randomly selected images, all resized to 299×299. For GPU measurements, a warm-up phase was performed by processing 10 images to stabilize GPU performance before timing the actual inference on 300 images. CPU and GPU inference times were then averaged per image and reported in milliseconds. All GPU experiments were conducted on a server equipped with an NVIDIA RTX 3090 GPU.
 
 To address the increased computational cost, future work will explore the use of lightweight backbone networks, particularly for the two frequency-domain branches, to reduce both parameter count and FLOPs, thereby improving the applicability of our network in real-world deployments.

\section{Conclusion}

This paper presents a novel deepfake detection network that jointly leverages spatial and frequency domain information via a triple-branch network. By analyzing individual frequency channels, we introduce a dynamic frequency channel selection module to capture a broader range of forgery artifacts, and a cross-frequency channel enhancement module to fuse primary and secondary frequency features. We further mathematically derive feature decoupling objectives based on mutual information theory to encourage feature diversity, and derive global information alignment objectives to effectively aggregate spatial and frequency features. 

Together, the \emph{triple-branch architecture}, the \emph{dynamic frequency channel selection}, and the \emph{mutual-information-based optimization} jointly constitute a progressively enhanced learning framework, in which the \emph{dynamic frequency channel selection} provides the primary performance gain, the \emph{triple-branch architecture} facilitates complementary spatial–frequency representation, and the \emph{mutual-information-based optimization} further stabilizes optimization and strengthens feature complementarity, ultimately yielding  cumulative improvements in both detection accuracy and cross-domain generalization.

Extensive experiments on six benchmark datasets demonstrate state-of-the-art performance in both in-dataset and cross-dataset scenarios. Our code is publicly available. Our future efforts will focus on improving computational efficiency and exploring domain adaptation to enhance real-world applicability.

\section*{Acknowledgment}

This research was supported by the National Natural Science Foundation of China (No.~62372402) and the Zhejiang Provincial Natural Science Foundation of China under Grant No.~LQN26F020059.

\bibliographystyle{IEEEtran}
\bibliography{ref}
\begin{IEEEbiography}[{\includegraphics[width=1in,height=1.25in,clip,keepaspectratio]{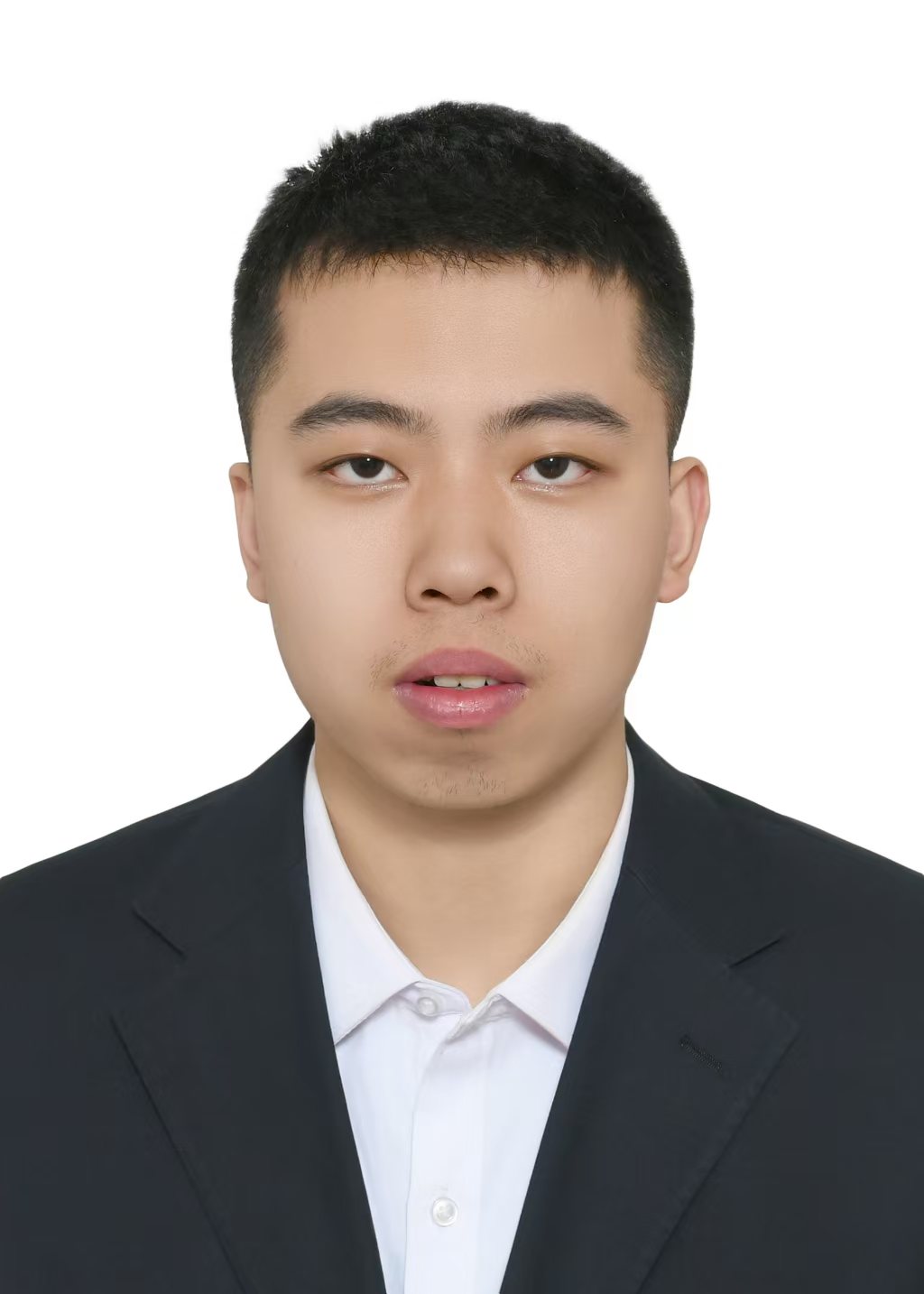}}]{Qihao Shen} received the B.S. degree in Software Engineering from Zhejiang University, Hangzhou, China, in 2024. He is currently pursuing the M.S. degree with the College of Engineering, Zhejiang University. His current research focuses on deepfake detection and multimedia forensics.
\end{IEEEbiography}

\begin{IEEEbiography}[{\includegraphics[width=1in,height=1.25in,clip,keepaspectratio]{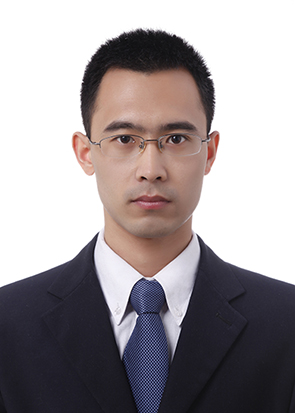}}]{Jiaxing Xuan} received the MS degree from Beijing University of Posts and Telecommunications in 2015. He currently works at State Grid Digital Technology Holding Co., Ltd., and is working toward the DEng degree with the College of Computer Science and Technology, Zhejiang University. His main research interests focus on the application of artificial intelligence technology in the energy and power industry.
\end{IEEEbiography}

\begin{IEEEbiography}[{\includegraphics[width=1in,height=1.25in,clip,keepaspectratio]{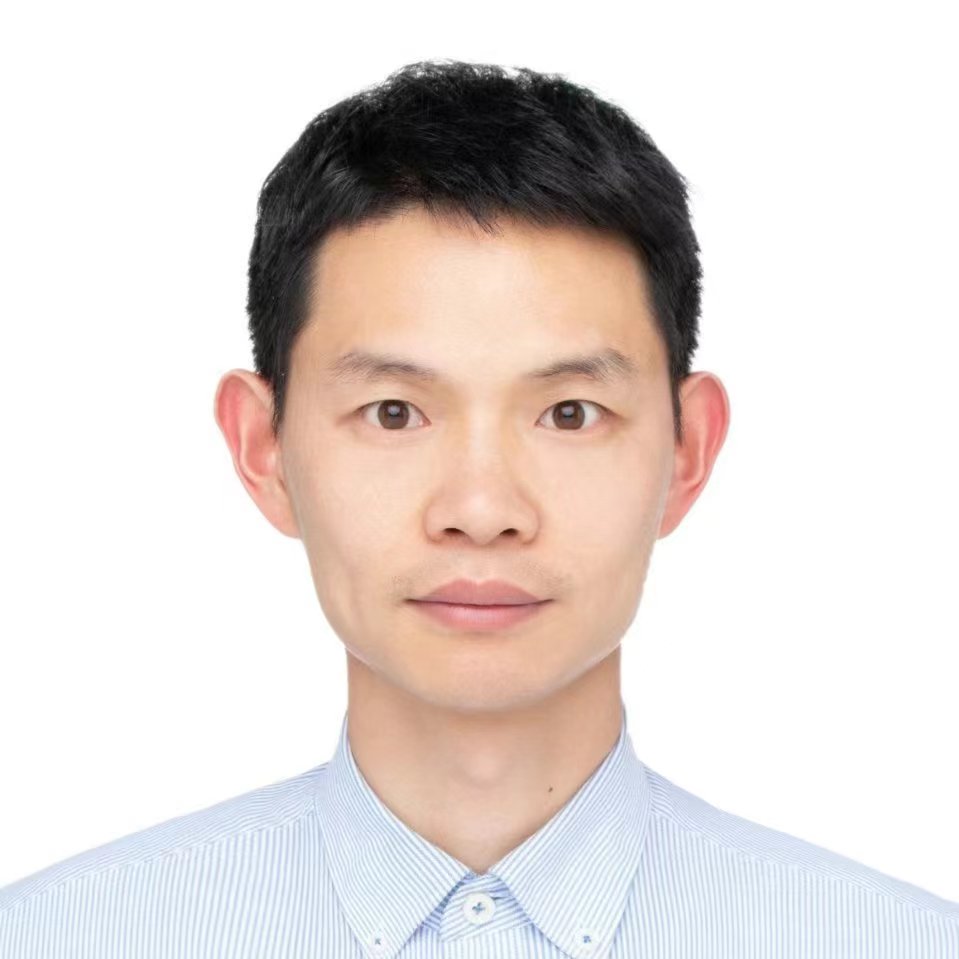}}]{Zhenguang Liu} is a professor of Zhejiang University. He had been a research fellow in National University of Singapore and A*STAR (Agency for Science, Technology and Research, Singapore). He respectively received his Ph.D. and B.E. degrees from Zhejiang University and Shandong University, China. His research interests include security and AI. Various parts of his work have been published in first-tier venues including PAMI, ACM CCS, S\&P, USENIX Security, TDSC, TIFS, CVPR, ICCV, TKDE, TIP, WWW, AAAI, ACM MM, ICLR, KDD, NeurlPS, INFOCOM, IJCAI, etc. Dr. Liu has served as technical program committee member for top-tier conferences such as CVPR, ICCV, WWW, AAAI, IJCAI, ACM MM, session chair of ICGIP, local chair of KSEM, and reviewer for IEEE TDSC, IEEE PAMI, IEEE TVCG, IEEE TIP, etc.
\end{IEEEbiography}

\begin{IEEEbiography}[{\includegraphics[width=1in,height=1.25in,clip,keepaspectratio]{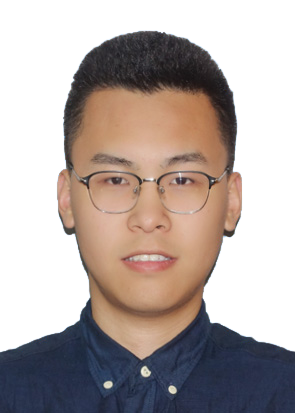}}]{Sifan Wu} is currently pursuing a Ph.D. degree at Jilin University. He received his B.E. and M.E. degrees from Hebei GEO University and Zhejiang Gongshang University, respectively. His research interests include pose estimation, motion copy, and deepfake detection.
\end{IEEEbiography}

\begin{IEEEbiography}[{\includegraphics[width=1in,height=1.25in,clip,keepaspectratio]{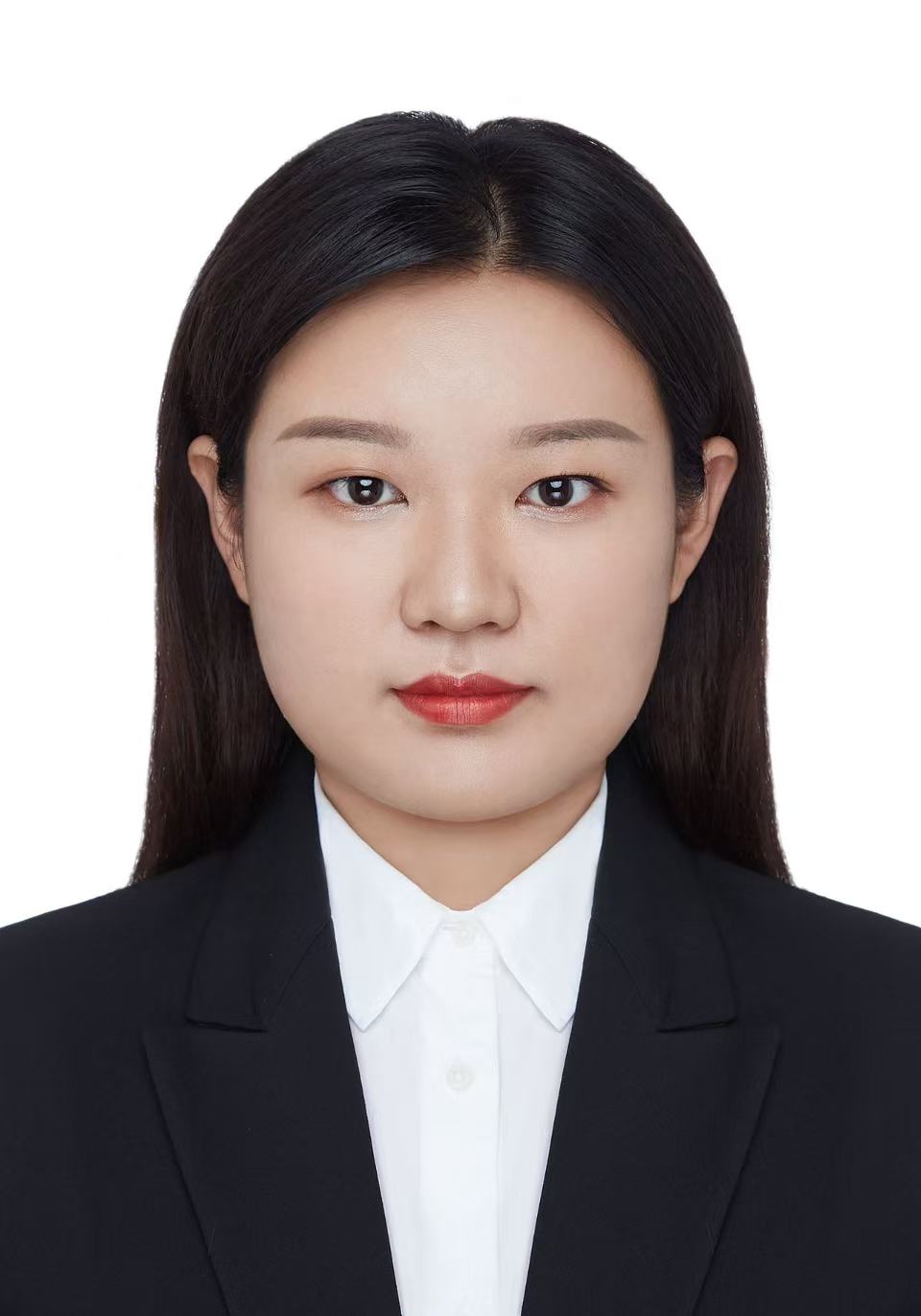}}]{Yutong Xie} received the M.S. degree in Software Engineering from Beijing Jiaotong University in 2021. She is currently a  Engineer at State Grid Blockchain Technology (Beijing) Co., Ltd. Her research interests include information security, artificial intelligence applications, data security and trustworthy intelligent systems.
\end{IEEEbiography}

\begin{IEEEbiography}[{\includegraphics[width=1in,height=1.25in,clip,keepaspectratio]{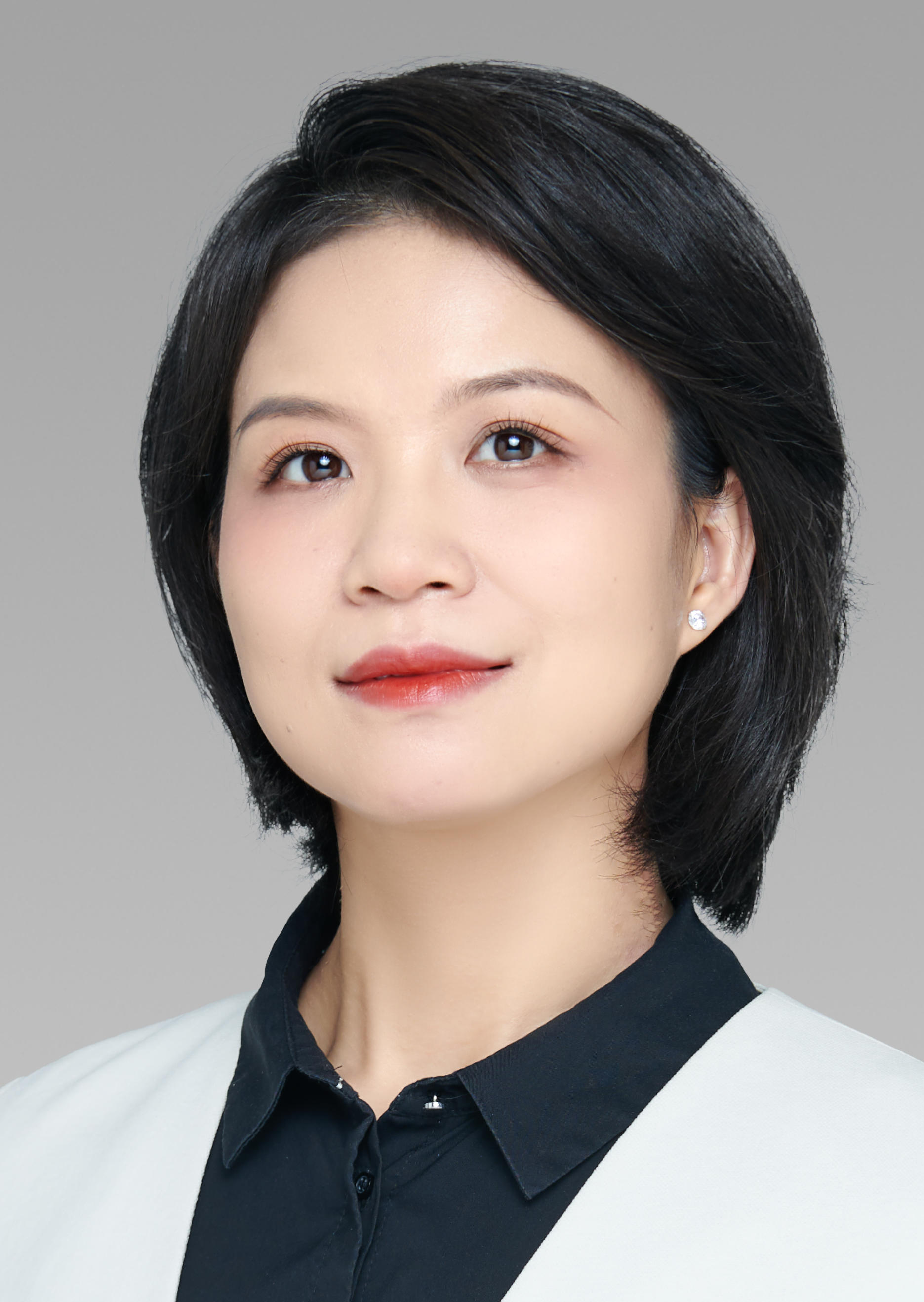}}]{Zhaoyan Ming} received the Ph.D. degree in computer science from the National University of Singapore, Singapore. She is currently an Associate Professor with Hangzhou City University, Hangzhou, China, where she specializes in AI for life and health science. She is actively involved in the academic communityas an Executive Member of the Technical Commit-tee on Natural Language Processing and the LargeLanguage Model Forum under the China ComputerFederation.
\end{IEEEbiography}

\begin{IEEEbiography}[{\includegraphics[width=1in,height=1.25in,clip,keepaspectratio]{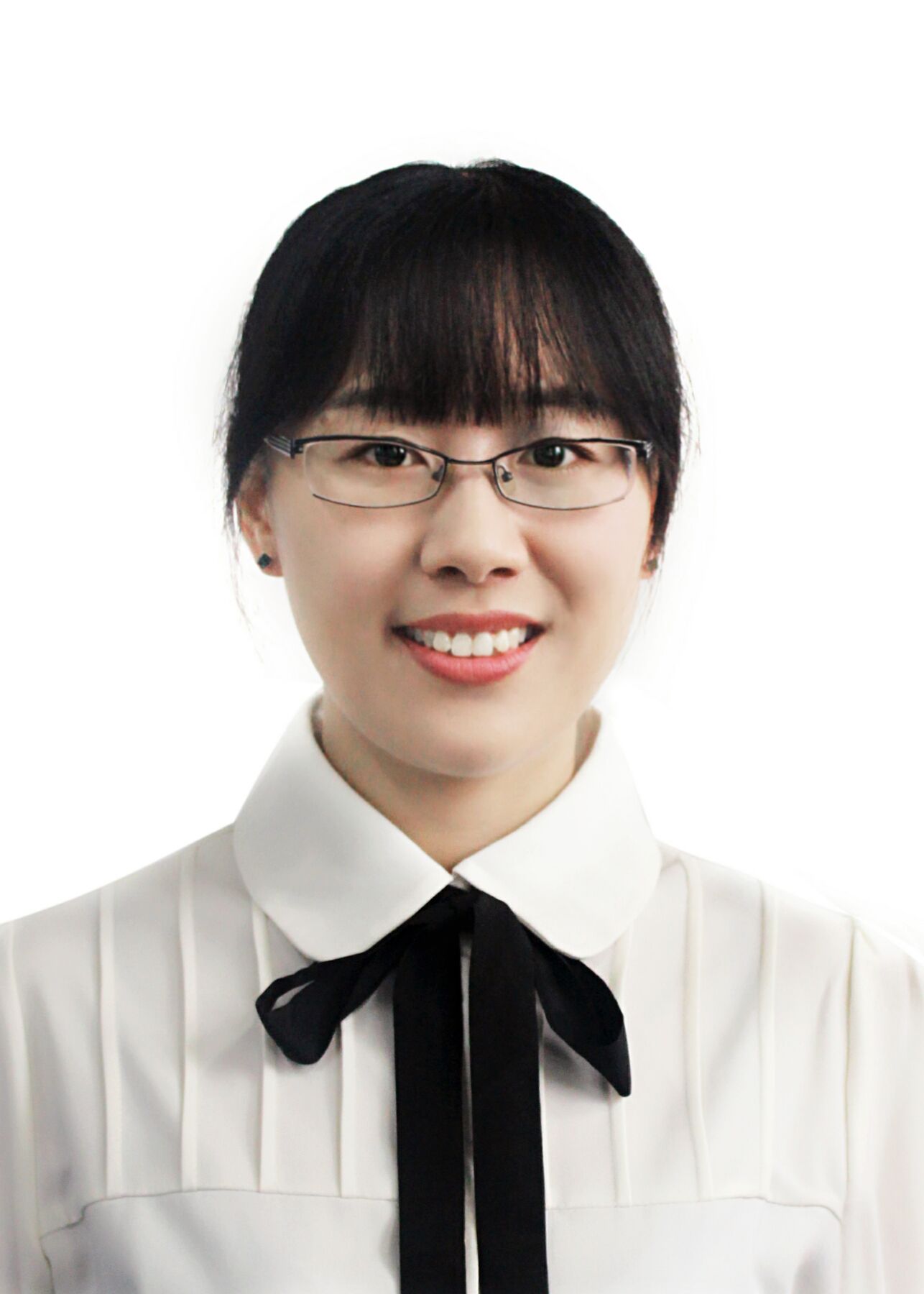}}]{Yingying Jiao} received the Ph.D. degree from Jilin University, Changchun, China, in 2025. She is currently an Associate Professor with the College of Computer Science and Technology, Zhejiang University of Technology, Hangzhou, China. She previously worked as a research assistant at the School of Computing, National University of Singapore. Her research interests encompass computer vision, multimedia processing, and security.
\end{IEEEbiography}

\begin{IEEEbiography}[{\includegraphics[width=1in,height=1.25in,clip,keepaspectratio]{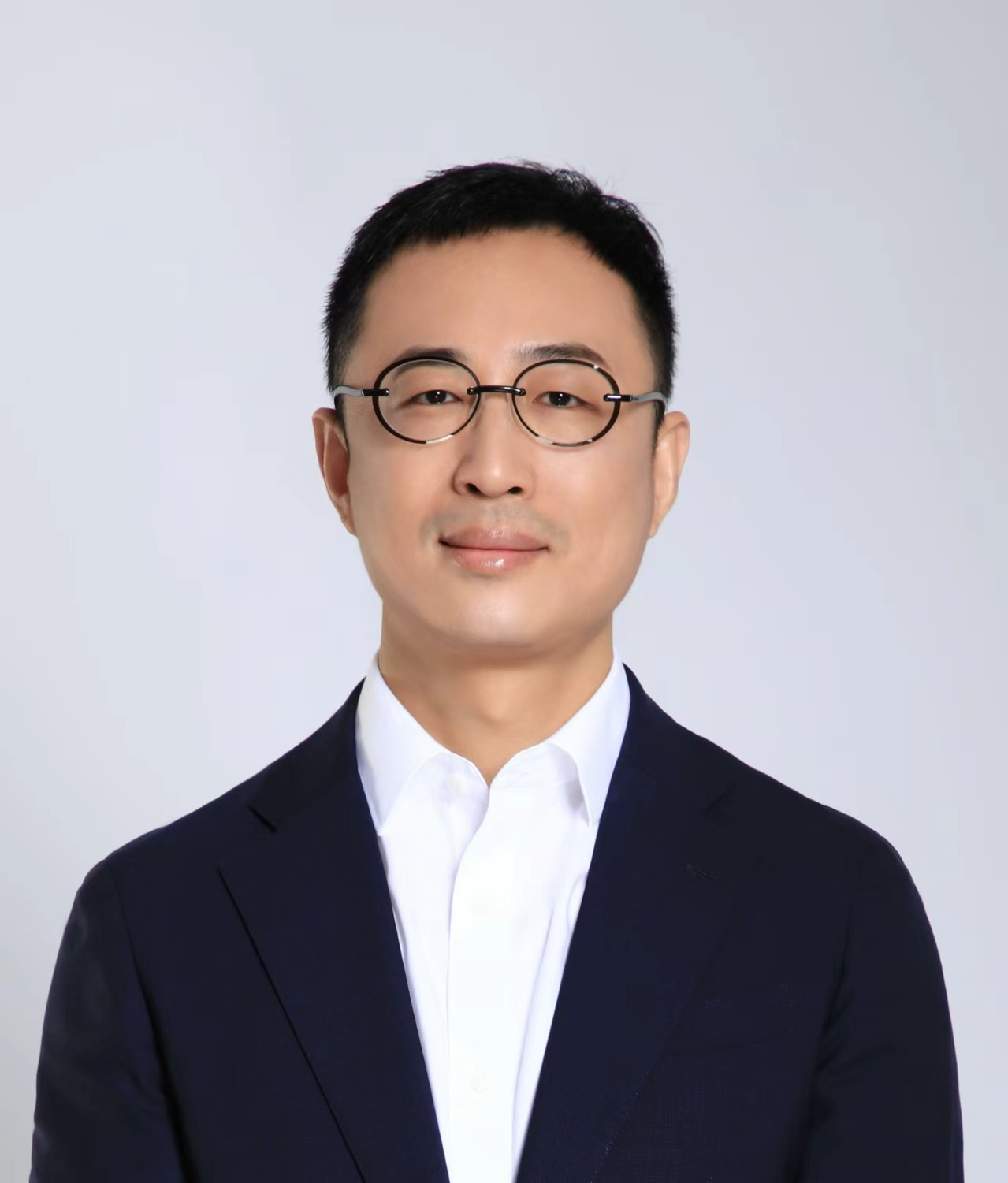}}]{Kui Ren}
 received degrees from three different majors, i.e., his Ph.D. in Electrical and Computer Engineering from Worcester Polytechnic Institute, USA, in 2007, M.Eng in Materials Engineering in 2001, and B.Eng in Chemical Engineering in 1998, both from Zhejiang University, China. Professor Kui Ren, AAAS, ACM, CCF, and IEEE Fellow, is currently the dean of the College of Computer Science and Technology at Zhejiang University. He is mainly engaged in research in data security and privacy protection, AI security, and computer applications.
\end{IEEEbiography}

\end{document}